\documentclass{article} 
\usepackage{iclr2027_conference,times}

\usepackage{amsmath,amsfonts,bm}

\def\eqref#1{equation~\ref{#1}}

\def\1{\bm{1}}

\DeclareMathAlphabet{\mathsfit}{\encodingdefault}{\sfdefault}{m}{sl}
\SetMathAlphabet{\mathsfit}{bold}{\encodingdefault}{\sfdefault}{bx}{n}

\usepackage{hyperref}
\usepackage{url}
\usepackage{graphicx}
\usepackage[font=small]{caption}
\usepackage{booktabs}
\usepackage{multirow}
\usepackage{array}
\usepackage{placeins}
\usepackage{amsmath}
\usepackage{flafter}
\usepackage{tikz}
\usetikzlibrary{positioning}
\usepackage[breakable,skins]{tcolorbox}
\usepackage{fvextra}
\usepackage{enumitem}
\definecolor{taskfill}{RGB}{243,243,240}
\definecolor{taskborder}{RGB}{115,115,115}
\definecolor{agentfill}{RGB}{235,244,252}
\definecolor{agentborder}{RGB}{0,114,178}
\definecolor{envfill}{RGB}{255,245,233}
\definecolor{envborder}{RGB}{213,94,0}
\newtcolorbox{taskbox}[1][]{colback=taskfill,colbacktitle=taskfill,
  colframe=taskborder,arc=3pt,boxrule=0.4mm,boxsep=0.8mm,breakable,
  fonttitle=\bfseries\small\color{black},fontupper=\small,left=1.2mm,right=1.2mm,
  title={#1}}
\newtcolorbox{agentturn}[1][]{colback=agentfill,colbacktitle=agentfill,
  colframe=agentborder,arc=3pt,boxrule=0.4mm,boxsep=0.8mm,breakable,
  fonttitle=\bfseries\small\color{black},fontupper=\small,left=1.2mm,right=1.2mm,
  title={Agent #1}}
\newtcolorbox{envturn}[1][]{colback=envfill,colbacktitle=envfill,
  colframe=envborder,arc=3pt,boxrule=0.4mm,boxsep=0.8mm,breakable,
  fonttitle=\bfseries\small\color{black},fontupper=\small,left=1.2mm,right=1.2mm,
  title={Terminal #1}}
\fvset{fontsize=\scriptsize,breaklines=true,breakanywhere=true,fontfamily=courier}

\newcommand{\actobs}{ActObs}
\newcommand{\actsft}{ActionSFT}
\newcommand{\obsact}{Obs$\rightarrow$Act}
\newcommand{\tbtwo}{Terminal-Bench~2.0}
\newcommand{\passk}{pass@$k$}
\newcommand{\gact}{g_{\mathrm{act}}}
\newcommand{\gobs}{g_{\mathrm{obs}}}
\newcommand{\pmse}[1]{{\fontsize{4.2pt}{4.8pt}\selectfont$\pm$#1}}

\title{{\fontsize{16pt}{20pt}\selectfont
Don't Mask the Environment: Observation Supervision Changes How Agents Explore Under RL}}

\author{
{\bfseries
Juzheng Zhang$^{1,2}$\thanks{Work done during an internship at AWS AI Labs.}\hspace{0.5em},\hspace{0.5em}
Disha Makhija$^{2}$\thanks{Equal contribution.}\hspace{0.4em},\hspace{0.5em} Manoj Ghuhan Arivazhagan$^{2}$\footnotemark[2]\hspace{0.4em},\hspace{0.5em}} \\
{\bfseries
\hspace{0.2em}Vinayshekhar Bannihatti Kumar$^{2}$\footnotemark[2]\hspace{0.4em},\hspace{0.5em} Rashmi Gangadharaiah$^{2}$} \\
$^{1}$University of Maryland \quad $^{2}$AWS AI Labs \\
{\texttt{juzheng@umd.edu} \hspace{0.3em}
\texttt{\{dismakhi,mghuhan,vinayshk,rgangad\}@amazon.com}}
}

\iclrfinalcopy
\begin{document}

\maketitle
\lhead{Preprint. Under review.}

\vspace{-4pt}
\begin{abstract}
Agent trajectories record what an agent does and what happens next. Yet standard supervised fine-tuning (SFT) applies loss only to agent-authored action tokens, using environment observations as context but not as prediction
targets. 
We ask whether this convention provides the best initialization for subsequent reinforcement learning. 
We introduce \actobs{}, which also supervises the observation tokens already present in each trajectory.
Although deployed agents never generate observations, learning to predict them encourages the policy to model action consequences without adding data, parameters, sequence tokens, or forward passes.
The methods perform similarly after SFT but diverge after GRPO. 
On Qwen3-4B, GRPO from \actobs{} achieves higher pass@$k$ at every evaluated sampling budget than its action-only counterpart on \tbtwo{}. 
On Qwen3-8B, it trades some pass@1 reliability for higher pass@$k$ (+3.4 pp at pass@16) and solves more distinct tasks. 
The advantage extends to cross-domain code editing on aider-polyglot (+4.2 pp at pass@1 at 4B), whose tasks are unseen during SFT and RL. 
\actobs{} retains more entropy during RL while requiring less policy movement,
leaving the final policy closer to its SFT initialization. 
Our analysis traces this difference to SFT: action and observation gradients rapidly become orthogonal, while action-only training leaves a large residual observation gradient and degrades environment prediction below the base model. 
Joint supervision prevents this one-sided specialization, preserving consequence prediction and preparing the policy for downstream exploration.
\end{abstract}

\section{Introduction}

A language-agent trajectory is more than a record of actions. It interleaves
decisions with their realized consequences: the agent issues a command, the
environment returns a new state, and the agent decides what to do next.
Nevertheless, standard trajectory SFT learns from only one side of this
interaction. The loss is applied to agent-authored actions, while environment
observations remain in the context but are excluded from the prediction targets
\citep{zeng2023agenttuning,chen2023fireact,chen2024agentflan}. This convention
appears natural because the deployed policy produces actions, not terminal
output. Masking observations assumes that
reading environmental feedback is useful, but learning to predict it is not.
That assumption has rarely been tested, especially when SFT is only the
initialization for subsequent RL.

\begin{figure}[t]
\centering
\resizebox{0.96\linewidth}{!}{%
\begin{tikzpicture}[
  every node/.style={font=\small},
  tok/.style={rectangle, rounded corners=1.2pt, minimum height=5mm, inner xsep=2.4pt, draw=black!45, line width=0.35pt},
  prompt/.style={tok, minimum width=13mm},
  action/.style={tok, minimum width=18mm},
  observation/.style={tok, minimum width=23mm},
  card/.style={rectangle, rounded corners=1.2pt, draw=black!35, line width=0.3pt,
    font=\fontsize{5.2}{6.2}\selectfont\ttfamily, align=left, anchor=south,
    inner xsep=1.5pt, inner ysep=1.8pt, minimum height=11mm},
  lab/.style={font=\small, anchor=east},
]
\node[prompt, fill=black!8] (x) {task $x$};
\node[action, fill=blue!14, right=1mm of x] (a1) {action $a_1$};
\node[observation, fill=orange!18, right=1mm of a1] (o1) {observation $o_1$};
\node[action, fill=blue!14, right=1mm of o1] (a2) {action $a_2$};
\node[observation, fill=orange!18, right=1mm of a2] (o2) {observation $o_2$};
\node[right=1mm of o2] (dots) {$\cdots$};
\node[lab, left=2.5mm of x] {Trajectory};
\node[card, fill=black!8, text width=26mm, anchor=south east]
  (cx) at ([yshift=2mm]x.north east)
  {You are an AI assistant solving command-line tasks [...]\\
   Task: start /app/alpine.iso in qemu so that telnet 127.0.0.1 6665
   reaches its login prompt.};
\node[card, fill=blue!14, text width=16.5mm, above=2mm of a1] (ca1)
  {"keystrokes": "qemu-system-\allowbreak x86\_64 -m 128M -cdrom
   /app/alpine.iso [...]"};
\node[card, fill=orange!18, text width=21.5mm, above=2mm of o1] (co1)
  {New Terminal Output:\\
   root@b6641d:/app\# qemu-system-x86\_64 -m 128M [...]\\
   {[1]} 131\\
   root@b6641d:/app\#};
\node[card, fill=blue!14, text width=16.5mm, above=2mm of a2] (ca2)
  {"keystrokes": "nc -z 127.0.0.1 6665 \&\& echo listening [...]"};
\node[card, fill=orange!18, text width=21.5mm, above=2mm of o2] (co2)
  {New Terminal Output:\\
   Port 6665 is listening\\
   root@b6641d:/app\#};
\foreach \c/\b in {cx/x, ca1/a1, co1/o1, ca2/a2, co2/o2}
  \draw[black!40, dashed, line width=0.3pt] (\c.south) -- (\b.north);
\node[prompt, fill=black!4, draw=black!25]
  (x2) at ([yshift=-8mm]x.center) {};
\node[action, fill=blue!35]
  (b1) at ([yshift=-8mm]a1.center) {$-\log p_\theta$};
\node[observation, fill=black!4, draw=black!25]
  (c1) at ([yshift=-8mm]o1.center) {masked};
\node[action, fill=blue!35]
  (b2) at ([yshift=-8mm]a2.center) {$-\log p_\theta$};
\node[observation, fill=black!4, draw=black!25]
  (c2) at ([yshift=-8mm]o2.center) {masked};
\node at ([yshift=-8mm]dots.center) {$\cdots$};
\node[lab, left=2.5mm of x2] {\actsft{}};
\node[prompt, fill=black!4, draw=black!25]
  (x3) at ([yshift=-16mm]x.center) {};
\node[action, fill=blue!35]
  (d1) at ([yshift=-16mm]a1.center) {$-\log p_\theta$};
\node[observation, fill=orange!45]
  (e1) at ([yshift=-16mm]o1.center) {$-\log p_\theta$};
\node[action, fill=blue!35]
  (d2) at ([yshift=-16mm]a2.center) {$-\log p_\theta$};
\node[observation, fill=orange!45]
  (e2) at ([yshift=-16mm]o2.center) {$-\log p_\theta$};
\node at ([yshift=-16mm]dots.center) {$\cdots$};
\node[lab, left=2.5mm of x3] {\actobs{}};
\end{tikzpicture}
}
\caption{\textbf{ActObs learns from the complete interaction.} The top row
shows a real agent trajectory (Appendix~\ref{app:transcripts}): a task,
the agent's actions, and the resulting terminal observations. Standard
\actsft{} masks the observations from the objective. \actobs{} trains on
the same sequence and exposes those already-available observation tokens to the loss.}
\label{fig:schematic}
\end{figure}
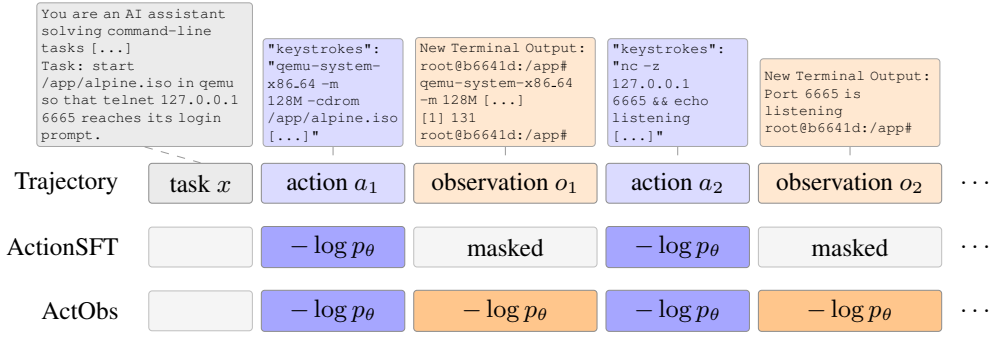

We study the counterintuitive alternative: train the policy to predict tokens
that it will never emit. As illustrated in Figure~\ref{fig:schematic},
\actobs{} simply unmasks the observations already present in each trajectory
and applies the language-modeling loss to both actions and observations. At an
observation position, this objective trains
$p_\theta(o_t\mid h_t,a_t)$, so the model must represent what the preceding
action does to the current environment. Each trajectory therefore serves as
both an imitation example and a transition example. Because the two objectives
share parameters, learning the action-to-consequence relation can shape the
representation used to choose later actions, an intuition shared by predictive
world-model approaches \citep{lin2023dynalang,guo2025worldmodelling,
zhang2025earlyexperience}. This additional constraint can preserve consequence
modeling and discourage one-sided specialization to action imitation.

We compare \actobs{} with \actsft{}, the standard action-only objective, and
\obsact{}, a timing control that applies observation-only SFT before
action-only SFT. The three SFT checkpoints perform similarly on \tbtwo{} but
diverge after the same GRPO procedure. At 4B, GRPO from \actobs{} yields the
strongest policy at every evaluated sampling budget,
including a 29\% relative advantage over \actsft{} at pass@1. At 8B, \actobs{}
gives up some single-attempt reliability but gains 14\% at pass@16
and solves 24 tasks rather than 21. The effect is even larger under
cross-domain transfer. On aider-polyglot's 225 multilingual code-editing
tasks, the 4B \actobs{} GRPO policy exceeds the ActionSFT GRPO policy by 43\% at
pass@1 and 24\% at pass@4, despite starting from a weaker code-editing
checkpoint before RL.

The entropy results point to a structured change in the learned policy. During
GRPO, \actobs{} maintains higher training entropy and ends with higher
self-entropy on its own evaluation rollouts, while remaining closer to its SFT
initialization. This retained uncertainty can keep a wider range of actions
accessible to repeated sampling, helping explain the stronger
pass@$k$ at larger $k$. Increasing the observation-loss weight reveals the
tradeoff: greater observation supervision progressively sacrifices
some pass@1 reliability while improving the probability of success under
repeated sampling. Yet raising the action-only policy's inference temperature 
to match the \actobs{} entropy does not close the gap in pass@$k$.

To understand why the same GRPO procedure produces different outcomes, 
we examine how the SFT objectives shape the initial policies for GRPO.
Action and observation gradients are initially aligned but rapidly 
become nearly orthogonal, so action-only updates no longer approximate 
the observation update. 
\actsft{} fits the action objective while leaving a large residual
observation gradient, and its ability to predict terminal feedback 
falls below that of the base model. 
\actobs{} continues to optimize the orthogonal component, reaching 
an SFT checkpoint that fits both objectives well. 
These distinct initializations shape subsequent GRPO training: 
the policy initialized from \actobs{} retains more entropy at 
positions specifying command arguments, flags, and paths, enabling 
repeated sampling to explore a wider range of command variants.

Our contributions are threefold:
\begin{itemize}[leftmargin=2em]
    \item \textbf{Objective.} We introduce \actobs{}, a simple change to the SFT loss mask that
    turns environment observations already present in expert trajectories into
    a predictive target without adding data, parameters, sequence tokens,
    forward passes, or changes to the RL algorithm.

    \item \textbf{Results.} Despite similar SFT performance, \actobs{} delivers stronger
    \passk{} performance and solves more distinct tasks after GRPO on
    \tbtwo{}, while transferring strongly to a cross-domain multilingual code-editing benchmark, 
    showing that SFT objectives shape
    downstream learning and exploration beyond immediate performance.

    \item \textbf{Mechanism.} We show that observation supervision changes the SFT solution
    and subsequent RL dynamics: it prevents one-sided gradient specialization,
    preserves environment prediction, and allows GRPO to retain more entropy
    with less policy movement. 
\end{itemize}

\section{ActObs: learning from the full agent trajectory}
\label{sec:method}

Every successful agent trajectory contains two aligned sources of
supervision: what the expert did and what the environment did next.
Standard agent SFT trains on the first while discarding the second as a
target, even though both streams are already present in the training
sequence. \actobs{} recovers this free signal with a single change to the
loss mask. The data, model, context, and training procedure remain fixed;
only the tokens included in the training loss are changed.

\paragraph{Objective.}
Let a trajectory be
$\tau=(x,a_1,o_1,a_2,o_2,\ldots,a_T,o_T)$, where $x$ is the task prompt,
$a_t$ is an assistant turn containing reasoning and commands, and $o_t$
is the response produced by the environment. In our setting, observations
are terminal outputs inserted verbatim into the next context. For the
tokenized trajectory $y_{1:N}$, let $\mathcal{A}$ and $\mathcal{O}$ denote
the action-token and observation-token indices. Standard agent SFT
optimizes only $\mathcal{A}$. \actobs{} instead minimizes
(Figure~\ref{fig:schematic})
\begin{equation}
\label{eq:actobs}
\mathcal{L}_{\lambda}(\theta)
= -\frac{
\displaystyle \sum_{i \in \mathcal{A}}
\log p_{\theta}(y_i \mid y_{<i})
+ \lambda \displaystyle \sum_{i \in \mathcal{O}}
\log p_{\theta}(y_i \mid y_{<i})
}{\lvert \mathcal{A} \rvert + \lambda \lvert \mathcal{O} \rvert}.
\end{equation}
$\lambda=0$ recovers action-only SFT (\actsft{}), while the default
\actobs{} uses $\lambda=1$. Prompts remain masked in both objectives. The
denominator preserves the per-example loss scale as $\lambda$ changes, and
at $\lambda=1$ action and observation tokens receive equal weight.
Because observations are already processed as context, ActObs requires no
new demonstrations, rollouts, parameters, or forward passes. In
implementation, it is only a label-mask change.

\paragraph{Consequence modeling.}
Although the deployed agent never generates observations, they remain a
valuable learning signal. Predicting $o_t$ requires anticipating what
$a_t$ will do to the environment. ActObs therefore turns each trajectory
into both a policy example and a transition example, training the same
representation that the next action must use.
Action-only SFT can sharpen the single
teacher command while overwriting pretrained knowledge of action
consequences, leaving useful alternatives with little probability. ActObs
fits decisions and outcomes jointly, preserving task-relevant probability
mass for subsequent RL. Appendix~\ref{app:ablations} reports controls for
sequential, observation-only, and shuffled-observation supervision.

\section{Main results}
\label{sec:experiments}

\subsection{Setup}
\label{sec:setup}

\begin{table}[t]
\caption{\textbf{Main results.} \passk{} in percent on \tbtwo{} (89 tasks; 16 attempts per
task) and aider-polyglot (225 code-editing tasks; four attempts per task).
ECHO combines GRPO with a next-observation prediction loss. 
Uncertainty is one bootstrap standard error, holding the task set fixed and
resampling attempts within each task. Boldface indicates the highest score within each stage.
The full \tbtwo{} results appear in
Appendix~\ref{app:fulltable}.}
\label{tab:main}
\begin{center}
\scriptsize
\setlength{\tabcolsep}{2.2pt}
\resizebox{\linewidth}{!}{%
\begin{tabular}{l cccccc c cccccc}
\toprule
& \multicolumn{6}{c}{Qwen3-4B} & & \multicolumn{6}{c}{Qwen3-8B} \\
\cmidrule{2-7} \cmidrule{9-14}
& \multicolumn{4}{c}{\tbtwo{}} & \multicolumn{2}{c}{aider-polyglot} & &
\multicolumn{4}{c}{\tbtwo{}} & \multicolumn{2}{c}{aider-polyglot} \\
\cmidrule(lr){2-5} \cmidrule(lr){6-7} \cmidrule(lr){9-12} \cmidrule(lr){13-14}
Model & @1 & @4 & @8 & @16 & @1 & @4 & & @1 & @4 & @8 & @16 & @1 & @4 \\
\midrule
\multicolumn{14}{l}{\emph{After SFT}} \\
Base & 1.2\pmse{0.2} & 3.3\pmse{0.5} & 4.7\pmse{0.6} & 5.6\pmse{0.7} & \textbf{3.0}\pmse{0.4} & \textbf{7.6}\pmse{0.7} & & 2.7\pmse{0.4} & 7.0\pmse{0.6} & 9.3\pmse{0.8} & 11.2\pmse{1.0} & 2.4\pmse{0.4} & 7.6\pmse{0.8} \\
\actsft{} & \textbf{4.5}\pmse{0.4} & \textbf{10.4}\pmse{0.8} & 14.0\pmse{1.0} & 16.9\pmse{1.2} & 1.4\pmse{0.3} & 5.8\pmse{0.7} & & \textbf{9.2}\pmse{0.5} & \textbf{18.2}\pmse{0.8} & \textbf{21.9}\pmse{1.0} & \textbf{24.7}\pmse{1.2} & 3.9\pmse{0.5} & 10.2\pmse{0.8} \\
\actobs{} & 4.4\pmse{0.4} & 9.9\pmse{0.9} & 14.0\pmse{1.1} & \textbf{18.0}\pmse{1.4} & 1.0\pmse{0.3} & 3.6\pmse{0.6} & & 7.9\pmse{0.5} & 16.1\pmse{0.8} & 19.7\pmse{0.9} & 22.5\pmse{1.2} & \textbf{4.4}\pmse{0.5} & \textbf{12.0}\pmse{0.9} \\
\obsact{} & 4.4\pmse{0.4} & 10.1\pmse{0.9} & \textbf{14.2}\pmse{1.1} & \textbf{18.0}\pmse{1.4} & 1.9\pmse{0.4} & 6.7\pmse{0.8} & & 8.3\pmse{0.5} & 15.9\pmse{0.8} & 19.8\pmse{1.1} & \textbf{24.7}\pmse{1.6} & 3.2\pmse{0.5} & 11.6\pmse{1.0} \\
\midrule
\multicolumn{14}{l}{\emph{After GRPO}} \\
\actsft{}$\rightarrow$GRPO & 5.6\pmse{0.4} & 11.1\pmse{0.7} & 14.0\pmse{1.0} & 18.0\pmse{1.4} & 9.7\pmse{0.6} & 20.4\pmse{0.9} & & \textbf{12.3}\pmse{0.5} & 18.7\pmse{0.6} & 21.0\pmse{0.8} & 23.6\pmse{1.2} & 12.7\pmse{0.8} & 28.4\pmse{1.3} \\
\actobs{}$\rightarrow$GRPO & \textbf{7.2}\pmse{0.4} & \textbf{12.5}\pmse{0.7} & \textbf{15.5}\pmse{1.0} & \textbf{19.1}\pmse{1.4} & \textbf{13.9}\pmse{0.7} & \textbf{25.3}\pmse{1.0} & & 11.0\pmse{0.5} & \textbf{19.3}\pmse{0.9} & \textbf{23.6}\pmse{1.1} & \textbf{27.0}\pmse{1.3} & \textbf{14.2}\pmse{0.8} & \textbf{28.9}\pmse{1.2} \\
\obsact{}$\rightarrow$GRPO & 5.3\pmse{0.4} & 11.4\pmse{0.8} & 14.7\pmse{0.9} & 16.9\pmse{1.0} & 10.0\pmse{0.7} & 21.3\pmse{0.9} & & 11.9\pmse{0.5} & \textbf{19.3}\pmse{0.7} & 21.7\pmse{0.7} & 23.6\pmse{1.0} & 12.6\pmse{0.7} & 24.4\pmse{1.1} \\
\midrule
\multicolumn{14}{l}{\emph{After ECHO}} \\
\actsft{}$\rightarrow$ECHO & \textbf{5.9}\pmse{0.4} & 11.6\pmse{0.8} & 14.7\pmse{1.0} & 18.0\pmse{1.3} & \textbf{10.2}\pmse{0.7} & 20.9\pmse{0.9} & & 9.6\pmse{0.5} & 16.5\pmse{0.9} & 20.6\pmse{1.2} & \textbf{25.8}\pmse{1.6} & \textbf{14.8}\pmse{0.8} & \textbf{29.8}\pmse{1.4} \\
\actobs{}$\rightarrow$ECHO & \textbf{5.9}\pmse{0.4} & \textbf{12.7}\pmse{0.8} & \textbf{16.5}\pmse{1.1} & \textbf{20.2}\pmse{1.4} & 9.1\pmse{0.6} & \textbf{21.8}\pmse{1.0} & & \textbf{11.4}\pmse{0.5} & \textbf{18.6}\pmse{0.8} & \textbf{22.0}\pmse{0.9} & 24.7\pmse{1.2} & 14.2\pmse{0.7} & 28.9\pmse{1.3} \\
\bottomrule
\end{tabular}%
}
\end{center}
\end{table}

\begin{figure}[t]
\centering
\includegraphics[width=0.85\linewidth]{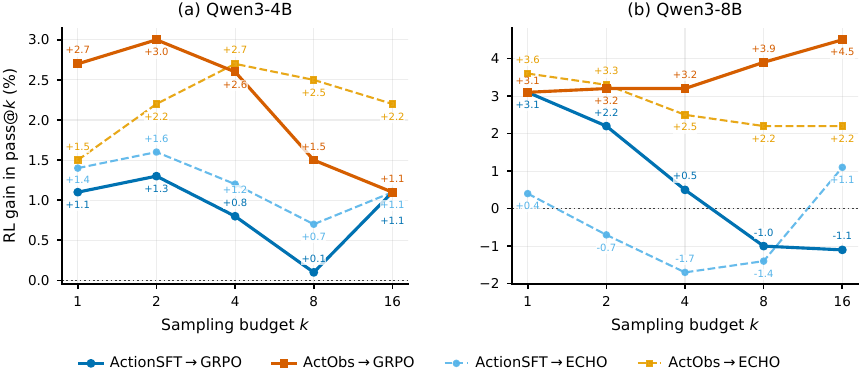}
\caption{\textbf{Performance gain during RL from each SFT initialization.}
Each curve reports $\Delta$\passk{}, the post-RL score minus the pre-RL SFT
score for the same initialization. Solid curves use standard GRPO.
Dashed curves use ECHO. Positive values indicate improvement
during RL. At 4B, GRPO yields larger gains from \actobs{} than from \actsft{} at every evaluated $k$. 
At 8B, the GRPO gains from \actobs{} increase with $k$, 
whereas those from \actsft{} are concentrated at small $k$ and turn negative for $k \geq 8$.
\actobs{} also yields larger ECHO gains than \actsft{} at every evaluated $k$ and model scale.}
\label{fig:delta}
\end{figure}

\paragraph{Training.} All methods share one corpus: 50k multi-turn terminal
trajectories (0.71B tokens) from the synthetic portion of
Nemotron-Terminal-Corpus \citep{nemotronterminal2026}, generated by
DeepSeek-V3.2.
Observation tokens are about 45\% of the corpus. We fine-tune Qwen3-4B
and Qwen3-8B \citep{qwen32025} for one epoch (781 steps, batch size 64,
cosine schedule, peak learning rate $10^{-5}$). The primary methods differ
only in their loss masks. We then apply GRPO \citep{shao2024deepseekmath}
for 135 steps on 2{,}392 containerized tasks from Endless Terminals
\citep{gandhi2026endless}, disjoint from the SFT corpus, using binary
verifiers (16 rollouts per task, batch size 32, learning rate $10^{-6}$).
Unless otherwise noted, observation supervision is used only during SFT;
standard GRPO contains no observation-prediction loss.
ECHO \citep{shrivastava2026echo} adds a next-observation loss
of weight 0.05 on the policy's own rollouts during GRPO.

\paragraph{Methods.} \actsft{} applies loss to actions only. \actobs{} uses
$\lambda=1$. \obsact{} is a timing control: one epoch of observation-only
SFT followed by one epoch of action-only SFT, inspired by work that develops
predictive environment models before downstream policy optimization
\citep{zhang2025earlyexperience,li2026word}.
It receives the same amount of observation supervision as \actobs{} over two epochs.

\paragraph{Evaluation.} All evaluation tasks are disjoint from the SFT
corpus and the Endless Terminals RL set. The primary benchmark is \tbtwo{}
\citep{terminalbench2025}, an OOD terminal benchmark with 89 tasks, the terminus-2 reference agent,
official wall-clock limits, and one pinned serving configuration for all
methods. Headline checkpoints are evaluated with 16
attempts per task. Model failures score zero; infrastructure
failures are rerun once. \passk{} uses the unbiased estimator of
\citet{chen2021codex}. Uncertainties are one bootstrap standard error,
computed by holding the task set fixed and resampling attempts within each
task.
Appendix~\ref{app:method} details the serving controls and statistical
procedure. Aider-polyglot tests cross-domain generalization on 225 unseen code-editing
tasks spanning six programming languages.

\subsection{Observation supervision strengthens downstream RL}
\label{sec:static}

Table~\ref{tab:main} summarizes our main results. After SFT, the three
initializations remain closely matched. After the same GRPO training, they
separate. Figure~\ref{fig:delta} reports the performance gain during
RL, while Table~\ref{tab:main} reports the absolute performance of the
resulting policies. Standard GRPO extracts a larger gain from \actobs{} at
every $k$ at 4B; at 8B, the gain from \actobs{} grows with $k$, whereas the
gain from \actsft{} becomes negative at large $k$.

At 4B, \actobs{} produces the strongest post-GRPO policy across the entire
sampling curve. It leads \actsft{}$\rightarrow$GRPO at pass@1, pass@4,
pass@8, and pass@16 by 1.6, 1.4, 1.5, and 1.1 percentage points,
corresponding to relative advantages of 29\%, 13\%, 11\%, and 6\%.
The effect therefore appears in single-attempt reliability and persists as
the sampling budget grows. At 8B, the advantage emerges at larger sampling
budgets.
\actsft{} leads at pass@1, but \actobs{} overtakes it by pass@4, and its
advantage grows from 0.6 points at pass@4 to 2.6 points at pass@8 and 3.4
points at pass@16, relative margins of 3\%, 12\%, and 14\%. \actobs{} reaches
a task coverage of 24, three tasks more than \actsft{}. It also solves three tasks that
no SFT policy or \actsft{}$\rightarrow$GRPO can solve, showing that RL from
the ActObs initialization reaches beyond the SFT-solvable set. The
sequential \obsact{} control does not
realize this benefit despite receiving the same amount of observation
supervision. At pass@16, \actobs{} leads the sequential control by 2.2 points
at 4B and 3.4 points at 8B. Joint action-observation
learning, rather than observation exposure alone, creates the stronger RL
initialization.

The advantage also survives when RL itself becomes observation-aware. ECHO
adds next-observation prediction during RL, yet \actobs{} leads or ties
\actsft{} at seven of the eight reported operating points. Relative to
\actsft{}$\rightarrow$ECHO, the \actobs{} lead ranges from 1.1 to 2.2 points
between pass@4 and pass@16 at 4B and from 1.4 to 2.1 points through pass@8
at 8B, relative margins as large as 12\% and 19\%, respectively. Only the
8B pass@16 comparison favors \actsft{}. Observation-aware RL therefore
complements rather than replaces the representation established by ActObs
during SFT, making \actobs{} a stronger initialization for ECHO than \actsft{}.

\subsection{Cross-domain transfer to code-editing tasks}
\label{sec:apoly}

We evaluate aider-polyglot as a cross-domain benchmark:
none of its 225 code-editing tasks appears in either the terminal SFT corpus
or the Endless Terminals RL set. Its six programming languages and
single-file editing objectives, evaluated by unit tests, define a task family
that is distinct from the environments used for training.
ActObs produces the strongest plain-GRPO policy at both model scales. At 4B,
it outperforms \actsft{}$\rightarrow$GRPO by 4.2 percentage points at pass@1
and 4.9 points at pass@4, relative advantages of 43\% and 24\%. It also exceeds
\obsact{}$\rightarrow$GRPO by 3.9 and 4.0 points. This advantage cannot be
attributed to a stronger code-editing policy before RL: the 4B ActObs SFT
checkpoint trails both alternatives on aider-polyglot. GRPO instead raises
ActObs by 12.9 points at pass@1 and 21.7 points at pass@4. 
At 8B, ActObs leads \actsft{}$\rightarrow$GRPO by 1.5 points
at pass@1, a 12\% relative advantage, and retains a 0.5-point lead at
pass@4. 
Among the plain-GRPO policies, it therefore has the strongest pass@1 and
pass@4.
Taken together, \tbtwo{} and aider-polyglot show that observation-aware SFT
can strengthen downstream RL on unseen terminal tasks and under cross-domain
transfer to multilingual code editing.

\section{ActObs preserves exploration during RL}
\label{sec:phenomenon}

Section~\ref{sec:experiments} shows that the SFT checkpoints have similar
benchmark performance but produce different policies after the same RL
procedure. We now trace where that difference appears. It is not visible in
the initial training reward, nor does it depend on reaching a higher final
training reward. Instead, the ActObs initialization changes the GRPO
trajectory, the amount of policy movement required by RL, and the uncertainty
retained by the final policy.

\subsection{Training dynamics during GRPO}
\label{sec:dynamics}

\begin{figure}[t]
\centering
\includegraphics[width=\linewidth]{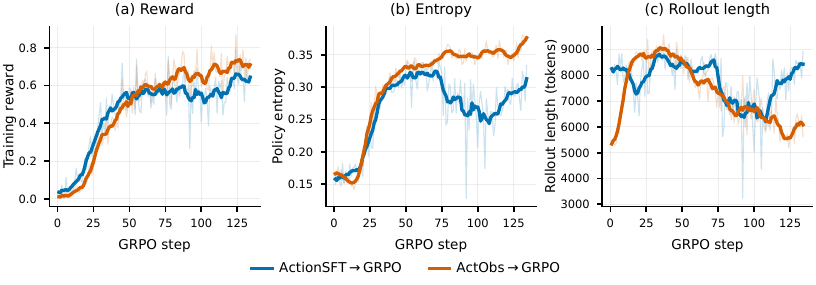}
\caption{\textbf{GRPO training dynamics at 4B.} (a) Training reward,
(b) on-policy training entropy, and (c) rollout length. The two runs begin
with similar reward and entropy, then diverge under the same GRPO procedure.
Compared with \actsft{}, \actobs{} finishes with higher reward, retains
increasing training entropy late in RL, and produces shorter final rollouts.
The corresponding 8B curves are shown in Figure~\ref{fig:dynamics8b}.}
\label{fig:dynamics}
\end{figure}

Figure~\ref{fig:dynamics} follows the quantities recorded online during GRPO.
The entropy in Figure~\ref{fig:dynamics}(b) is \emph{training entropy}: the
token-level policy entropy on the current on-policy rollout batch at each
update. The 4B runs begin with similar reward and training entropy, but their
trajectories separate. ActObs reward rises more smoothly and finishes higher,
while its training entropy continues to rise late in RL as ActionSFT entropy
falls and only partially recovers. Rollout length separates as well
(Figure~\ref{fig:dynamics}(c)). After the early phase, ActObs rollouts become
progressively shorter and remain shorter at the end, whereas ActionSFT rollout
length turns upward late in training. Thus, the higher late-stage entropy of
ActObs does not come from generating longer trajectories.

The same qualitative entropy separation appears at 8B: ActObs maintains
higher training entropy after the curves diverge, even though the runs converge
in training reward (Figure~\ref{fig:dynamics8b}).
Across scales, the initialization changes how the
policy evolves under the same GRPO objective rather than simply providing more
reward headroom.

\subsection{Self-entropy at the final checkpoints}
\label{sec:entropy}
\label{sec:evalpolicy}

\begin{figure}[t]
\centering
\includegraphics[width=0.9\linewidth]{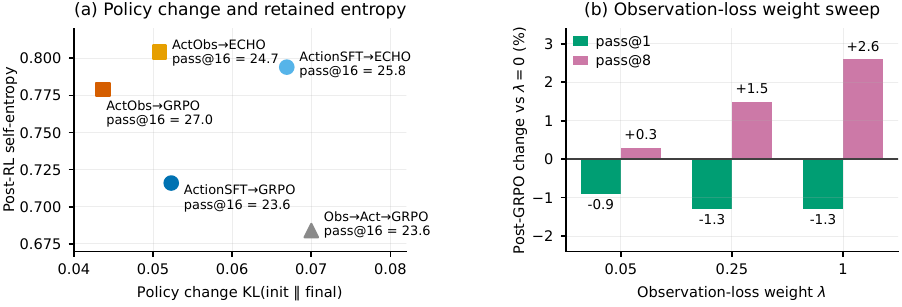}
\caption{\textbf{Observation supervision controls the post-RL policy at 8B.}
(a) Endpoint KL versus self-entropy for each post-RL policy. KL is measured
from its SFT initialization on 200 shared \tbtwo{} traces, self-entropy on
200 of the policy's own \tbtwo{} evaluation rollouts, and labels report pass@16.
\actobs{}$\rightarrow$GRPO combines the smallest endpoint displacement with
high retained entropy and the highest pass@16. (b) Post-GRPO
pass@1 and pass@8 relative to the action-only endpoint ($\lambda=0$) as the
observation-loss weight increases. Stronger observation supervision
progressively shifts performance from pass@1 reliability toward
pass@8 coverage.}
\label{fig:mechlam}
\end{figure}

The vertical axis of Figure~\ref{fig:mechlam}(a) reports \emph{self-entropy}.
To compute it, we freeze each RL endpoint, select 200 rollouts from that
policy's \tbtwo{} evaluation, and average its next-token entropy over
the assistant tokens. Self-entropy therefore measures the final policy's
uncertainty on states that the policy itself visits. This differs from the
\emph{training entropy} in Figure~\ref{fig:dynamics}(b), which is recorded on
the newly sampled on-policy training batch at every GRPO update. 
Appendix~\ref{app:onpolicy} also reports
\emph{fixed-state entropy}: every checkpoint scores the same 200 rollout
traces, holding the inputs fixed to control for state visitation.

ActObs retains more post-RL self-entropy than ActionSFT while reaching higher pass@16
(Figure~\ref{fig:mechlam}(a)). The fixed-state probe in
Appendix~\ref{app:onpolicy} shows the same ordering. ActObs therefore leaves
non-negligible probability on a wider set of plausible command continuations.
With multiple rollouts, these commands are more likely to be sampled,
raising \passk{} as $k$ grows.

\phantomsection
\label{sec:temp}
Higher entropy by itself is not sufficient. ECHO produces the highest
endpoint entropy but does not match \actobs{}$\rightarrow$GRPO at pass@16. We also raise
the sampling temperature of \actsft{}$\rightarrow$GRPO until its self-entropy
matches that of \actobs{}$\rightarrow$GRPO under the same self-entropy probe
used above ($T^{*}=0.64$ instead of $0.6$). Across 16 attempts,
temperature matching changes pass@1, pass@4, pass@8, and pass@16 by at most 0.7
points, solves no additional task, and leaves the pass@16 gap intact. The
high-$k$ advantage therefore comes from the distribution learned by the
policy, not from injecting more randomness at inference time.

\subsection{Less policy movement}
\label{sec:kl}

The horizontal axis of Figure~\ref{fig:mechlam}(a) compares each final RL
policy with the SFT checkpoint from which it was initialized. For every pair,
we score both policies at action-token positions on the same fixed set of 200
\tbtwo{} traces and average
$\mathrm{KL}(\pi_{\mathrm{init}}\,\|\,\pi_{\mathrm{final}})$. The states are
therefore identical across all pairs, and the value measures endpoint
displacement. 
ActObs undergoes the smallest endpoint change, retains more
self-entropy, and attains the strongest pass@16.
ActionSFT moves farther and retains less entropy, while \obsact{}
moves farthest and retains the least.

This joint ordering explains how smaller policy movement can preserve higher
entropy. Because the ActObs endpoint remains closer to its own initialization,
GRPO reallocates less probability mass and leaves more of the starting
distribution within reach of repeated sampling. Prior work similarly
finds that RLVR can improve pass@1 by concentrating probability on rewarded
paths while reducing the set of problems solved at large $k$
\citep{yue2025rlvrlimit,wu2025invisibleleash}. ActObs limits this contraction,
preserving more of the initial policy support for repeated sampling. ECHO
marks the boundary of this explanation:
it produces still higher entropy but does not match ActObs at pass@16.
Therefore, high entropy alone is not enough. ActObs combines limited policy
movement with high retained entropy and higher \passk{} at larger $k$.

\subsection{Varying observation supervision strength}
\label{sec:dose}

Recall that $\lambda$ in Eq.~\ref{eq:actobs} controls the weight on
observation-token prediction: $\lambda=0$ is action-only SFT, and
$\lambda=1$ is the default ActObs objective. Figure~\ref{fig:mechlam}(b)
varies this weight while holding the data and downstream GRPO recipe fixed.
As $\lambda$ increases, the post-GRPO pass@8 advantage over action-only SFT
grows monotonically, while pass@1 moves in the opposite direction. The sweep
therefore reveals a pass@1-to-pass@8 tradeoff rather than a uniform
shift in performance. Lower observation weight favors success from a single
rollout; higher observation weight sacrifices some pass@1 but increases the
chance that at least one of several rollouts succeeds. Because the SFT
checkpoints remain similar in benchmark performance, $\lambda$ controls how
the initialization allocates probability mass for later RL, with its benefit
emerging at larger sampling budgets.

\section{Why joint supervision is a stronger initialization}
\label{sec:mechanism}

\subsection{Gradient dynamics during SFT}
\label{sec:imprint}

\begin{figure}[t]
\centering
\includegraphics[width=\linewidth]{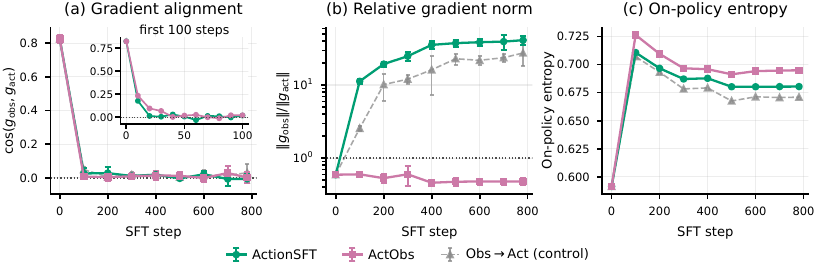}
\caption{\textbf{Joint supervision prevents one-sided gradient specialization.} 
At each saved SFT checkpoint, action and observation gradients are
computed separately on the same 256 held-out trajectories. Error bars 
show standard deviation across four equal partitions. (a) Cosine
similarity $c$ starts at 0.83 and falls to the empirical noise floor within 10
to 20 SFT steps. (b) Norm ratio $r$ grows to about 41 under \actsft{} but stays
near 0.5 under \actobs{}. (c) On-policy entropy measured on 200 \tbtwo{}
evaluation rollouts per method. The separation appears by step 100 and persists through
the rest of SFT.}
\label{fig:imprint}
\end{figure}

The observation loss cannot directly constrain standard GRPO because it is
absent from the RL objective. Its effect is instead carried by the SFT
checkpoint delivered to GRPO. Action-only SFT fits the demonstrated decisions
while allowing the model's prediction of their environmental consequences to
deteriorate. Joint supervision keeps both signals active through the SFT-to-RL
handoff and therefore presents RL with a different policy to refine.

We measure how this separation develops using checkpoints saved throughout
the \actsft{} and \actobs{} SFT runs. At every checkpoint, we separately compute
the action-token and observation-token gradients on the same 256 held-out
trajectories. Thus both gradients are measured for both methods.
For \actsft{}, the observation gradient is a diagnostic only; it was masked
during training and is never applied to the checkpoints.

Let $\gact$ and $\gobs$ denote the summed action-token and observation-token
gradients. We use the two quantities plotted in Figure~\ref{fig:imprint}
throughout: their cosine similarity
$c=\cos(\gobs,\gact)$ and norm ratio
$r=\lVert\gobs\rVert/\lVert\gact\rVert$. Let
$\hat{g}_{\mathrm{act}}=\gact/\lVert\gact\rVert$ be the unit vector along the
action-gradient direction, and let $\hat{g}_{\mathrm{act}}^{\perp}$ be the
unit vector along the component of $\gobs$ orthogonal to $\gact$. We can then write
\begin{equation}
\label{eq:decomp}
\gobs=r\lVert\gact\rVert\left(c\hat{g}_{\mathrm{act}}+\sqrt{1-c^{2}}\,\hat{g}_{\mathrm{act}}^{\perp}\right).
\end{equation}
This form maps directly onto the figure. Figure~\ref{fig:imprint}(a) reports $c$, which controls
how the observation gradient divides between the parallel and orthogonal
directions. Figure~\ref{fig:imprint}(b) reports its overall scale $r$. When $c\approx0$, the
parallel fraction disappears, $\sqrt{1-c^{2}}\approx1$, and $r$ directly
approximates the size of the orthogonal observation component relative to the
action gradient.

At the pretrained checkpoint in Figure~\ref{fig:imprint}(a), the two token streams initially
favor a largely shared update. Within 10 to 20 SFT steps, $c$ falls to the
noise floor and remains there. The parallel fraction of the observation
gradient has therefore disappeared, leaving an almost entirely orthogonal
signal. Action-only training cannot supply the orthogonal direction, while ActObs continues
to receive this component through its observation loss.
Figure~\ref{fig:imprint}(b) shows the magnitude of that orthogonal signal. Under \actsft{}, $r$
rises to about 41 because fitting the demonstrated actions makes $\gact$
small while the unfitted $\gobs$ remains large. Under \actobs{}, both signals
are optimized and $r$ stays near 0.5 through the final checkpoint. This
orthogonal update is what prevents one-sided specialization: after the shared
direction disappears, ActObs continues to reduce the observation residual
along a direction that action-only training never follows. The two objectives
therefore end SFT in different regions of the joint loss landscape. ActionSFT
is nearly stationary for the action loss but remains steep in an orthogonal
observation direction; ActObs keeps the two signals balanced.
Functionally, ActObs preserves the base model's observation prediction, while
action-only fitting erodes it.

Figure~\ref{fig:imprint}(c) tracks on-policy entropy across SFT checkpoints,
measured on 200 \tbtwo{} evaluation rollouts per method. The entropy curves separate by
step 100 of 781 and remain apart through the end of SFT. Joint
supervision thus establishes the policy difference early and maintains it
after the initial gradient alignment has disappeared.

\subsection{Token-level effects}
\label{sec:where}

\begin{figure}[t]
\centering
\includegraphics[width=\linewidth]{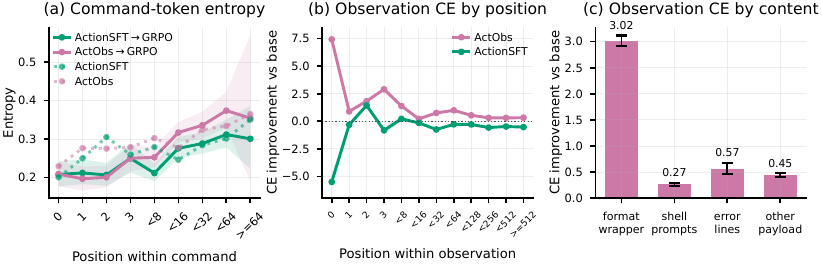}
\caption{\textbf{Where observation supervision changes the policy and the
prediction target.} (a) Command-token entropy by position on 200 of each
checkpoint's own \tbtwo{} evaluation rollouts (solid: RL checkpoints; dotted:
SFT checkpoints; shading: bootstrap 95\% intervals over traces). After GRPO,
\actobs{} and \actsft{} are similarly sharp at the first command tokens, but
\actobs{} retains more entropy later in the command. (b,c) Teacher-forced
observation cross-entropy on a shared set of 300 held-out trajectories from
the Nemotron-Terminal-Corpus validation split, reported as improvement over
the base model. (b) groups tokens by position, and (c) groups tokens by content type.}
\label{fig:position}
\end{figure}

Every terminal observation in the SFT data begins with the fixed wrapper
``New Terminal Output:'', whose short repeated prefix occupies the first few
observation positions. This regular prefix is an unusually easy prediction
target and must be distinguished from the variable terminal payload. The
scalar entropy measurements in Section~\ref{sec:entropy} do not reveal where
the additional uncertainty lies. Figure~\ref{fig:position}(a) shows that the
difference is highly structured. We freeze each SFT or RL checkpoint, score
its next-token distributions on 200 of its own \tbtwo{} evaluation rollouts,
and bucket command-token entropy
by position. After GRPO, the first few command tokens are similarly sharp
under \actobs{} and \actsft{}, and the separation opens later in the command,
where tokens commonly encode arguments, flags, paths, and other choices.
ActObs therefore does not simply make the agent less decisive at the beginning
of a tool call. It preserves more variation in how a selected command is
instantiated. A representative \actobs{}-only success uses this freedom to
sweep QEMU launch parameters one at a time until it finds a working memory
configuration (Appendix~\ref{app:transcripts}).

Figures~\ref{fig:position}(b) and~\ref{fig:position}(c) locate the training
signal that accompanies this policy difference. Here the base model,
\actsft{}, and \actobs{} are teacher-forced on a shared set of 300 held-out
Nemotron-Terminal-Corpus validation trajectories, and each ground-truth
observation token is scored after its preceding trajectory context. The
plotted quantity is the reduction in cross-entropy relative to the base model,
so positive values mean better prediction. As expected, the fixed wrapper
produces the largest early cross-entropy gain, but it is not the whole effect:
\actobs{} also improves at later positions and on shell prompts, error lines,
and other payload text, whereas \actsft{} is worse than the base model across
most of the sequence. Observation supervision therefore preserves prediction
of environmental feedback following actions beyond memorizing the response
header. By contrast, ActionSFT degrades observation prediction below the base
model, suggesting that its weaker downstream RL performance may partly stem
from losing this capacity.

\section{Related work}
\label{sec:related}

\paragraph{Trajectory supervision and agent RL.}
Language-agent SFT distills multi-turn interaction traces from human or
stronger-model demonstrations
\citep{zeng2023agenttuning,chen2023fireact,chen2024agentflan,pan2024swegym}.
Although these traces interleave agent actions with environment feedback,
representative pipelines compute loss only on assistant outputs, leaving
observations as context rather than prediction targets
\citep{zeng2023agenttuning,chen2024agentflan}. Prior work mainly
improves how trajectories are collected, filtered, and formatted, including at
the scale of terminal-agent corpora \citep{nemotronterminal2026}. \actobs{}
targets this overlooked choice: holding the teacher, trajectories, and token
sequence fixed, it extends supervision from action tokens to both action and
observation tokens. Agent RL instead learns directly from interaction rewards,
with recent work improving optimization for multi-turn agents and building
scalable environments with verifiable feedback
\citep{wang2025ragen,da2025agentrlvr,gandhi2026endless,ivison2026tmax}. We keep
the GRPO recipe unchanged and isolate the initialization supplied by SFT.

\paragraph{World modeling for language agents.}
Predictive objectives and world models have long supported representation
learning and imagined control
\citep{jaderberg2017unreal,pathak2017curiosity,ha2018worldmodels}.
For language agents, predicted outcomes guide planning
\citep{lin2023dynalang,qiao2024wkm,gu2024webdreamer,chae2025webworld} or
self-verification \citep{guo2025worldmodelling}. During RL, ECHO \citep{shrivastava2026echo}, 
PaW \citep{lu2026cotrain}, and TAPO \citep{li2026tapo}
reuse on-policy transitions for observation supervision, while RWML rewards
agreement between predicted and realized states \citep{yu2026rwml}; AAWM uses
agent-authored, decision-oriented targets instead of literal next observations
\citep{cai2026beyondobs}. Closest to our setting, Early Experience collects new
transitions before sequential world-model SFT, imitation, and GRPO
\citep{zhang2025earlyexperience}; SPA inserts explicit state descriptions
through self-play \citep{chen2025internalizing}; and From Word to World \citep{li2026word} and
Qwen-AgentWorld \citep{qwenagentworld2026} use world-model training as a separate policy warm-up. 
\actobs{} requires none of these extra
transitions, targets, stages, or inference-time simulators. It only unmasks raw
environment observations already present in the same expert traces, letting us
isolate how the SFT loss mask shapes subsequent GRPO.

\paragraph{Initialization and exploration in RLVR.}
SFT can stabilize the behavior needed for RL, yet excessive SFT can reduce
plasticity at the handoff \citep{deepseekr1,chu2025sftmemorizes,
liu2026plasticity}. Exploration-aware fine-tuning and posterior behavioral
cloning further show that similar pre-RL performance can conceal different
initial policy distributions or action support
\citep{mu2026explorationaware,wagenmaker2025posteriorbc}. During RLVR, entropy
often falls as reward rises, and higher pass@1 can come with fewer rare
solutions under repeated sampling
\citep{cui2025entropy,yu2025dapo,yue2025rlvrlimit,wu2025invisibleleash}; ProRL
shows that prolonged, KL-controlled training can instead expand this boundary
\citep{liu2025prorl}. Existing methods mainly intervene during RL through
clipping, entropy-aware updates, or \passk{} objectives
\citep{yu2025dapo,wang2025highentropy,cheng2025explorationentropy,
chen2025passktraining}. \actobs{} intervenes earlier by changing the checkpoint
handed to otherwise unchanged RL.

\section{Conclusion}
\label{sec:conclusion}

We asked whether an agent should be supervised only on the tokens it will
later emit.
\actobs{} learns from both the teacher's actions and the environment's 
responses, using information already present in every trajectory. Predicting
the feedback that follows an action encourages the shared model to represent
how that action changes the environment, shaping the policy that later chooses
actions. This change barely affects benchmark performance immediately after SFT, yet it produces a
substantially stronger starting point for GRPO. At 4B, \actobs{} achieves higher
\passk{} at every evaluated sampling budget after GRPO. At 8B, it trades some single-shot
reliability for greater success under repeated sampling. It also transfers
strongly to a cross-domain multilingual code-editing benchmark, showing that the benefit extends
beyond the terminal tasks. Our analysis traces these gains to
the SFT geometry. As action and observation gradients become orthogonal,
action-only SFT fits actions while leaving a large residual observation gradient
and degrading environment prediction. Joint supervision avoids this one-sided
specialization, enabling GRPO to retain more entropy with less policy movement.

\begingroup
\small
\bibliographystyle{iclr2027_conference}
\bibliography{references_verified}

@article{shrivastava2026echo,
  title        = {{ECHO}: Terminal Agents Learn World Models for Free},
  author       = {Shrivastava, Vaishnavi and Kauffmann, Piero and Awadallah, Ahmed and Papailiopoulos, Dimitris},
  journal      = {arXiv preprint arXiv:2605.24517},
  year         = {2026}
}

@article{shao2024deepseekmath,
  title        = {{DeepSeekMath}: Pushing the Limits of Mathematical Reasoning in Open Language Models},
  author       = {Shao, Zhihong and Wang, Peiyi and Zhu, Qihao and Xu, Runxin and Song, Junxiao and Bi, Xiao and Zhang, Haowei and Zhang, Mingchuan and Li, Y. K. and Wu, Y. and others},
  journal      = {arXiv preprint arXiv:2402.03300},
  year         = {2024}
}

@article{yu2025dapo,
  title        = {{DAPO}: An Open-Source {LLM} Reinforcement Learning System at Scale},
  author       = {Yu, Qiying and Zhang, Zheng and Zhu, Ruofei and Yuan, Yufeng and Zuo, Xiaochen and Yue, Yu and Dai, Weinan and Fan, Tiantian and Liu, Gaohong and Liu, Lingjun and others},
  journal      = {arXiv preprint arXiv:2503.14476},
  year         = {2025}
}

@article{cui2025entropy,
  title        = {The Entropy Mechanism of Reinforcement Learning for Reasoning Language Models},
  author       = {Cui, Ganqu and Zhang, Yuchen and Chen, Jiacheng and Yuan, Lifan and Wang, Zhi and Zuo, Yuxin and Li, Haozhan and Fan, Yuchen and Chen, Huayu and Chen, Weize and others},
  journal      = {arXiv preprint arXiv:2505.22617},
  year         = {2025}
}

@article{wang2025highentropy,
  title        = {Beyond the 80/20 Rule: High-Entropy Minority Tokens Drive Effective Reinforcement Learning for {LLM} Reasoning},
  author       = {Wang, Shenzhi and Yu, Le and Gao, Chang and Zheng, Chujie and Liu, Shixuan and Lu, Rui and Dang, Kai and Chen, Xionghui and Yang, Jianxin and Zhang, Zhenru and others},
  journal      = {arXiv preprint arXiv:2506.01939},
  year         = {2025}
}

@inproceedings{jaderberg2017unreal,
  title        = {Reinforcement Learning with Unsupervised Auxiliary Tasks},
  author       = {Jaderberg, Max and Mnih, Volodymyr and Czarnecki, Wojciech Marian and Schaul, Tom and Leibo, Joel Z and Silver, David and Kavukcuoglu, Koray},
  booktitle    = {International Conference on Learning Representations (ICLR)},
  year         = {2017}
}

@inproceedings{ha2018worldmodels,
  title        = {Recurrent World Models Facilitate Policy Evolution},
  author       = {Ha, David and Schmidhuber, J\"urgen},
  booktitle    = {Advances in Neural Information Processing Systems (NeurIPS)},
  year         = {2018}
}

@inproceedings{pathak2017curiosity,
  title        = {Curiosity-Driven Exploration by Self-Supervised Prediction},
  author       = {Pathak, Deepak and Agrawal, Pulkit and Efros, Alexei A and Darrell, Trevor},
  booktitle    = {International Conference on Machine Learning (ICML)},
  year         = {2017}
}

@inproceedings{terminalbench2025,
  title        = {{Terminal-Bench}: Benchmarking Agents on Hard, Realistic Tasks in Command Line Interfaces},
  author       = {Merrill, Mike A. and Shaw, Alexander G. and Carlini, Nicholas and Li, Boxuan and Raj, Harsh and Bercovich, Ivan and Shi, Lin and Shin, Jeong Yeon and Walshe, Thomas and Buchanan, E. Kelly and others},
  booktitle    = {International Conference on Learning Representations (ICLR)},
  year         = {2026},
  eprint       = {2601.11868},
  archivePrefix = {arXiv}
}

@article{qwen32025,
  title        = {{Qwen3} Technical Report},
  author       = {Yang, An and Li, Anfeng and Yang, Baosong and Zhang, Beichen and Hui, Binyuan and Zheng, Bo and Yu, Bowen and Gao, Chang and Huang, Chengen and Lv, Chenxu and others},
  journal      = {arXiv preprint arXiv:2505.09388},
  year         = {2025}
}

@article{nemotronterminal2026,
  title={On Data Engineering for Scaling {LLM} Terminal Capabilities},
  author={Pi, Renjie and Lam, Grace and Shoeybi, Mohammad and Jannaty, Pooya and Catanzaro, Bryan and Ping, Wei},
  journal={arXiv preprint arXiv:2602.21193},
  year={2026}
}

@article{qwenagentworld2026,
  title={{Qwen-AgentWorld}: Language World Models for General Agents},
  author={Zuo, Yuxin and Xiao, Zikai and Sheng, Li and Huang, Fei and Tu, Jianhong and Liu, Yuxuan and Tang, Tianyi and Hu, Xiaomeng and Su, Yang and Lan, Qingfeng and others},
  journal={arXiv preprint arXiv:2606.24597},
  year={2026}
}

@inproceedings{lin2023dynalang,
  title={Learning to Model the World with Language},
  author={Lin, Jessy and Du, Yuqing and Watkins, Olivia and Hafner, Danijar and Abbeel, Pieter and Klein, Dan and Dragan, Anca},
  booktitle={International Conference on Machine Learning (ICML)},
  year={2024},
  eprint={2308.01399},
  archivePrefix={arXiv}
}

@article{qiao2024wkm,
  title={Agent Planning with World Knowledge Model},
  author={Qiao, Shuofei and Fang, Runnan and Zhang, Ningyu and Zhu, Yuqi and Chen, Xiang and Deng, Shumin and Jiang, Yong and Xie, Pengjun and Huang, Fei and Chen, Huajun},
  journal={arXiv preprint arXiv:2405.14205}, year={2024}
}

@article{gu2024webdreamer,
  title={Is Your LLM Secretly a World Model of the Internet? Model-Based Planning for Web Agents},
  author={Gu, Yu and Zhang, Kai and Ning, Yuting and Zheng, Boyuan and Gou, Boyu and Xue, Tianci and Chang, Cheng and Srivastava, Sanjari and Xie, Yanan and Qi, Peng and others},
  journal={arXiv preprint arXiv:2411.06559}, year={2024}
}

@article{lu2026cotrain,
  title={Policy and World Modeling Co-Training for Language Agents},
  author={Lu, Ning and Lin, Baijiong and Liu, Shengcai and Wu, Jiahao and Lv, Haoze and Wei, Yanbin and Zhu, Lingting and Qian, Shengju and Wang, Xin and Chen, Ying-Cong and others},
  journal={arXiv preprint arXiv:2606.02388}, year={2026}
}

@article{cai2026beyondobs,
  title={Beyond Next-Observation Prediction: Agent-Authored World Modeling for Sequential Decision Making},
  author={Cai, Guangfeng and Yang, Kaibing and He, Shuo and Li, Yu and Yang, Shengtian and Lv, Jiaqi and Feng, Lei},
  journal={arXiv preprint arXiv:2606.25421}, year={2026}
}

@article{zeng2023agenttuning,
  title={AgentTuning: Enabling Generalized Agent Abilities for LLMs},
  author={Zeng, Aohan and Liu, Mingdao and Lu, Rui and Wang, Bowen and Liu, Xiao and Dong, Yuxiao and Tang, Jie},
  journal={arXiv preprint arXiv:2310.12823}, year={2023}
}

@article{chen2024agentflan,
  title={Agent-FLAN: Designing Data and Methods of Effective Agent Tuning for Large Language Models},
  author={Chen, Zehui and Liu, Kuikun and Wang, Qiuchen and Zhang, Wenwei and Liu, Jiangning and Lin, Dahua and Chen, Kai and Zhao, Feng},
  journal={arXiv preprint arXiv:2403.12881}, year={2024}
}

@article{cheng2025explorationentropy,
  title={Reasoning with Exploration: An Entropy Perspective},
  author={Cheng, Daixuan and Huang, Shaohan and Zhu, Xuekai and Dai, Bo and Zhao, Wayne Xin and Zhang, Zhenliang and Wei, Furu},
  journal={arXiv preprint arXiv:2506.14758}, year={2025}
}

@article{yue2025rlvrlimit,
  title={Does Reinforcement Learning Really Incentivize Reasoning Capacity in LLMs Beyond the Base Model?},
  author={Yue, Yang and Chen, Zhiqi and Lu, Rui and Zhao, Andrew and Wang, Zhaokai and Yue, Yang and Song, Shiji and Huang, Gao},
  journal={arXiv preprint arXiv:2504.13837}, year={2025}
}

@article{chen2025passktraining,
  title={Pass@k Training for Adaptively Balancing Exploration and Exploitation of Large Reasoning Models},
  author={Chen, Zhipeng and Qin, Xiaobo and Wu, Youbin and Ling, Yue and Ye, Qinghao and Zhao, Wayne Xin and Shi, Guang},
  journal={arXiv preprint arXiv:2508.10751}, year={2025}
}

@article{liu2025prorl,
  title={ProRL: Prolonged Reinforcement Learning Expands Reasoning Boundaries in Large Language Models},
  author={Liu, Mingjie and Diao, Shizhe and Lu, Ximing and Hu, Jian and Dong, Xin and Choi, Yejin and Kautz, Jan and Dong, Yi},
  journal={arXiv preprint arXiv:2505.24864}, year={2025}
}

@article{wu2025invisibleleash,
  title={The Invisible Leash: Why RLVR May or May Not Escape Its Origin},
  author={Wu, Fang and Xuan, Weihao and Lu, Ximing and Liu, Mingjie and Dong, Yi and Harchaoui, Zaid and Choi, Yejin},
  journal={arXiv preprint arXiv:2507.14843}, year={2025}
}

@article{chu2025sftmemorizes,
  title={SFT Memorizes, RL Generalizes: A Comparative Study of Foundation Model Post-training},
  author={Chu, Tianzhe and Zhai, Yuexiang and Yang, Jihan and Tong, Shengbang and Xie, Saining and Schuurmans, Dale and Le, Quoc V. and Levine, Sergey and Ma, Yi},
  journal={arXiv preprint arXiv:2501.17161}, year={2025}
}

@article{liu2026plasticity,
  title={When RL Fails after SFT: Rejuvenating Model Plasticity for Robust SFT-to-RL Handoff},
  author={Liu, Runze and Liu, Jiashun and Wan, Xu and Fu, Yuqian and Pan, Ling},
  journal={arXiv preprint arXiv:2606.09932}, year={2026}
}

@article{mu2026explorationaware,
  title={Offline Exploration-Aware Fine-Tuning for Long-Chain Mathematical Reasoning},
  author={Mu, Yongyu and Zeng, Jiali and Meng, Fandong and Zhu, JingBo and Xiao, Tong},
  journal={arXiv preprint arXiv:2603.16206}, year={2026}
}

@article{chen2021codex,
  title        = {Evaluating Large Language Models Trained on Code},
  author       = {Chen, Mark and Tworek, Jerry and Jun, Heewoo and Yuan, Qiming and Pinto, Henrique Ponde de Oliveira and Kaplan, Jared and Edwards, Harri and Burda, Yuri and Joseph, Nicholas and Brockman, Greg and others},
  journal      = {arXiv preprint arXiv:2107.03374},
  year         = {2021}
}

@article{deepseekr1,
  title        = {{DeepSeek-R1}: Incentivizing Reasoning Capability in {LLMs} via Reinforcement Learning},
  author       = {{DeepSeek-AI}},
  journal      = {arXiv preprint arXiv:2501.12948},
  year         = {2025}
}

@article{chen2023fireact,
  title        = {{FireAct}: Toward Language Agent Fine-tuning},
  author       = {Chen, Baian and Shu, Chang and Shareghi, Ehsan and Collier, Nigel and Narasimhan, Karthik and Yao, Shunyu},
  journal      = {arXiv preprint arXiv:2310.05915},
  year         = {2023}
}

@article{pan2024swegym,
  title        = {Training Software Engineering Agents and Verifiers with {SWE-Gym}},
  author       = {Pan, Jiayi and Wang, Xingyao and Neubig, Graham and Jaitly, Navdeep and Ji, Heng and Suhr, Alane and Zhang, Yizhe},
  journal      = {arXiv preprint arXiv:2412.21139},
  year         = {2024}
}

@article{gandhi2026endless,
  title        = {{Endless Terminals}: Scaling {RL} Environments for Terminal Agents},
  author       = {Gandhi, Kanishk and Garg, Shivam and Goodman, Noah D. and Papailiopoulos, Dimitris},
  journal      = {arXiv preprint arXiv:2601.16443},
  year         = {2026}
}

@article{ivison2026tmax,
  title        = {{Tmax}: A Simple Recipe for Terminal Agents},
  author       = {Ivison, Hamish and Yin, Junjie Oscar and Shao, Rulin and Xiao, Teng and Lambert, Nathan and Hajishirzi, Hannaneh},
  journal      = {arXiv preprint arXiv:2606.23321},
  year         = {2026}
}

@article{guo2025worldmodelling,
  title        = {World Modelling Improves Language Model Agents},
  author       = {Guo, Shangmin and Darwiche Domingues, Omar and Avalos, Rapha{\"e}l and Courville, Aaron and Strub, Florian},
  journal      = {arXiv preprint arXiv:2506.02918},
  year         = {2025}
}

@article{zhang2025earlyexperience,
  title        = {Agent Learning via Early Experience},
  author       = {Zhang, Kai and Chen, Xiangchao and Liu, Bo and Xue, Tianci and Liao, Zeyi and Liu, Zhihan and Wang, Xiyao and Ning, Yuting and Chen, Zhaorun and Fu, Xiaohan and others},
  journal      = {arXiv preprint arXiv:2510.08558},
  year         = {2025},
  note         = {ICML 2026}
}

@article{chen2025internalizing,
  title        = {Internalizing World Models via Self-Play Finetuning for Agentic {RL}},
  author       = {Chen, Shiqi and Zhu, Tongyao and Wang, Zian and Zhang, Jinghan and Wang, Kangrui and Gao, Siyang and Xiao, Teng and Teh, Yee Whye and He, Junxian and Li, Manling},
  journal      = {arXiv preprint arXiv:2510.15047},
  year         = {2025}
}

@inproceedings{li2026word,
  title        = {From Word to World: Can Large Language Models be Implicit Text-based World Models?},
  author       = {Li, Yixia and Wang, Hongru and Qiu, Jiahao and Yin, Zhenfei and Zhang, Dongdong and Qian, Cheng and Li, Zeping and Ma, Xiaoteng and Chen, Guanhua and Ji, Heng},
  booktitle    = {Proceedings of the 64th Annual Meeting of the Association for Computational Linguistics (Volume 1: Long Papers)},
  pages        = {8084--8111},
  publisher    = {Association for Computational Linguistics},
  year         = {2026},
  doi          = {10.18653/v1/2026.acl-long.366}
}

@article{yu2026rwml,
  title        = {Reinforcement World Model Learning for {LLM}-based Agents},
  author       = {Yu, Xiao and Peng, Baolin and Xu, Ruize and Shen, Yelong and He, Pengcheng and Nath, Suman and Singh, Nikhil and Gao, Jiangfeng and Yu, Zhou},
  journal      = {arXiv preprint arXiv:2602.05842},
  year         = {2026}
}

@article{li2026tapo,
  title        = {{TAPO}: Transition-Aware Policy Optimization for {LLM} Agents},
  author       = {Li, Cong and Peng, Peixi and Zhao, Yisen and Hu, Xinyu and Liu, Shudong and Su, Zhan and Li, Zhuojian},
  journal      = {arXiv preprint arXiv:2607.27973},
  year         = {2026}
}

@inproceedings{chae2025webworld,
  title        = {Web Agents with World Models: Learning and Leveraging Environment Dynamics in Web Navigation},
  author       = {Chae, Hyungjoo and Kim, Namyoung and Ong, Kai Tzu-iunn and Gwak, Minju and Song, Gwanwoo and Kim, Jihoon and Kim, Sunghwan and Lee, Dongha and Yeo, Jinyoung},
  booktitle    = {International Conference on Learning Representations (ICLR)},
  year         = {2025}
}

@article{wang2025ragen,
  title        = {{RAGEN}: Understanding Self-Evolution in {LLM} Agents via Multi-Turn Reinforcement Learning},
  author       = {Wang, Zihan and Wang, Kangrui and Wang, Qineng and Zhang, Pingyue and Li, Linjie and Yang, Zhengyuan and Jin, Xing and Yu, Kefan and Nguyen, Minh Nhat and Liu, Licheng and others},
  journal      = {arXiv preprint arXiv:2504.20073},
  year         = {2025}
}

@article{da2025agentrlvr,
  title        = {{Agent-RLVR}: Training Software Engineering Agents via Guidance and Environment Rewards},
  author       = {Da, Jeff and Wang, Clinton and Deng, Xiang and Ma, Yuntao and Barhate, Nikhil and Hendryx, Sean},
  journal      = {arXiv preprint arXiv:2506.11425},
  year         = {2025}
}

@article{wagenmaker2025posteriorbc,
  title        = {Posterior Behavioral Cloning: Pretraining {BC} Policies for Efficient {RL} Finetuning},
  author       = {Wagenmaker, Andrew and Dong, Perry and Tsao, Raymond and Finn, Chelsea and Levine, Sergey},
  journal      = {arXiv preprint arXiv:2512.16911},
  year         = {2025}
}
\endgroup

\newpage
\appendix
\section{Evaluation protocol and statistical procedures}
\label{app:method}

\paragraph{Reporting convention.}
We evaluate all 89 \tbtwo{} tasks and all 225 aider-polyglot tasks. Here, $n$
denotes the number of attempts per task. The
headline \tbtwo{} results pool four independently launched batches of four
attempts per task, giving each evaluated SFT and RL checkpoint 16 attempts per
task. Other evaluations use four or eight attempts per task as indicated.
Model-caused failures, including formatting failures, receive a score of zero.
A timeout can still count as a success if the verifier confirms that the task
was completed. Infrastructure failures from the harness, verifier, or Docker
are rerun once under the same configuration. We compute \passk{} with the
unbiased estimator of \citet{chen2021codex}.

\paragraph{Uncertainty estimates.}
Tables report the point estimate and one bootstrap standard error, computed
from 20{,}000 resamples. Each bootstrap replicate holds the benchmark task
set fixed, samples the observed binary attempts with replacement within each
task, recomputes the \passk{} estimator, and averages it across
tasks. The resulting uncertainty captures finite-attempt sampling variation.

\paragraph{Evaluation budget.}
Using 16 attempts per task for every headline comparison 
reduces sampling variance in the \passk{} estimates, 
enabling a more reliable comparison in 
Table~\ref{tab:main}. All headline checkpoints use the same task set, attempt
count, and serving configuration, so each comparison has a matched evaluation
budget. The same samples support estimates from pass@1 through pass@16 without
changing the evaluation protocol across sampling budgets.

\paragraph{Serving configuration.}
\tbtwo{} task limits are defined in wall-clock time, making serving latency
part of the evaluation protocol. We pinned the serving setup after a
nine-setting sweep over tensor parallelism, model colocation, and trial
concurrency. For \tbtwo{} evaluation, a single vLLM engine uses tensor
parallelism across eight NVIDIA A100 GPUs on one node, colocated with the
Docker trials. The setup allows 12 concurrent trials, a 40{,}960-token
context, and vLLM GPU-memory utilization of 0.90.
Decoding follows the checkpoint's generation configuration:
$T=0.6$, top-$p$ 0.95, and top-$k$ 20, except for the explicit
temperature-matching analysis in Section~\ref{sec:temp}.

\section{Complete results and ablations}
\label{app:fulltable}
\label{app:ablations}

\paragraph{Complete 8B matrix and control definitions.}
\begin{table}[t]
\caption{\textbf{Complete Qwen3-8B results on \tbtwo{}.} The results use the pinned
serving configuration and the reporting protocol in Appendix~\ref{app:method}.
Rows above the horizontal rule report SFT checkpoints, and rows below it
report RL checkpoints. Each checkpoint in the primary comparison is evaluated
with 16 attempts per task.}
\label{tab:full8b}
\begin{center}
\footnotesize
\begin{tabular}{l c ccccc c}
\toprule
Model & $n$ & pass@1 & pass@2 & pass@4 & pass@8 & pass@16 & solved/89 \\
\midrule
Base (no SFT) & 16 & 2.7\pmse{0.4} & 4.6\pmse{0.5} & 7.0\pmse{0.6} & 9.3\pmse{0.8} & 11.2\pmse{1.0} & 10 \\
\actsft{} & 16 & 9.2\pmse{0.5} & 13.6\pmse{0.7} & \textbf{18.2}\pmse{0.8} & \textbf{21.9}\pmse{1.0} & \textbf{24.7}\pmse{1.2} & \textbf{22} \\
\actobs{} & 16 & 7.9\pmse{0.5} & 11.8\pmse{0.7} & 16.1\pmse{0.8} & 19.7\pmse{0.9} & 22.5\pmse{1.2} & 20 \\
\obsact{} & 16 & 8.3\pmse{0.5} & 12.0\pmse{0.6} & 15.9\pmse{0.8} & 19.8\pmse{1.1} & \textbf{24.7}\pmse{1.6} & \textbf{22} \\
act2act (2 epochs action) & 16 & \textbf{9.3}\pmse{0.5} & 13.2\pmse{0.6} & 17.0\pmse{0.8} & 20.7\pmse{1.1} & \textbf{24.7}\pmse{1.4} & \textbf{22} \\
\actobs{}-100 (first 100 steps) & 16 & 8.4\pmse{0.5} & 12.6\pmse{0.7} & 17.0\pmse{0.8} & 20.9\pmse{0.9} & 23.6\pmse{1.1} & 21 \\
$\lambda=0.05$ & 4 & 9.0\pmse{0.5} & 13.3\pmse{0.7} & 18.0\pmse{0.8} & -- & -- & -- \\
$\lambda=0.25$ & 4 & \textbf{9.3}\pmse{0.6} & \textbf{13.9}\pmse{0.7} & 18.0\pmse{0.8} & -- & -- & -- \\
$\lambda=0.5$ & 4 & 7.9\pmse{0.7} & 12.0\pmse{0.8} & 16.9\pmse{0.9} & -- & -- & -- \\
$\lambda=2.0$ & 4 & 8.1\pmse{0.5} & 11.4\pmse{0.6} & 14.6\pmse{0.8} & -- & -- & -- \\
Obs only & 4 & 0.0 & 0.0 & 0.0 & -- & -- & -- \\
Shuffled obs & 4 & 3.7\pmse{0.7} & 6.0\pmse{0.9} & 9.0\pmse{1.2} & -- & -- & -- \\
Wild logs & 4 & 7.9\pmse{0.9} & 12.2\pmse{1.2} & 16.9\pmse{1.5} & -- & -- & -- \\
Salient-filtered obs & 4 & 7.2\pmse{0.8} & 10.3\pmse{1.0} & 13.5\pmse{1.3} & -- & -- & -- \\
Salient \obsact{} & 4 & 8.1\pmse{0.9} & 12.4\pmse{1.1} & 15.7\pmse{1.2} & -- & -- & -- \\
Action$\rightarrow$obs-only & 4 & 0.0 & 0.0 & 0.0 & -- & -- & -- \\
\midrule
\actsft{}$\rightarrow$GRPO & 16 & \textbf{12.3}\pmse{0.5} & \textbf{15.9}\pmse{0.5} & 18.7\pmse{0.6} & 21.0\pmse{0.8} & 23.6\pmse{1.2} & 21 \\
\actobs{}$\rightarrow$GRPO & 16 & 11.0\pmse{0.5} & 15.0\pmse{0.7} & 19.3\pmse{0.9} & \textbf{23.6}\pmse{1.1} & \textbf{27.0}\pmse{1.3} & \textbf{24} \\
\obsact{}$\rightarrow$GRPO & 16 & 11.9\pmse{0.5} & \textbf{15.9}\pmse{0.6} & 19.3\pmse{0.7} & 21.7\pmse{0.7} & 23.6\pmse{1.0} & 21 \\
\actsft{}$\rightarrow$ECHO & 16 & 9.6\pmse{0.5} & 13.0\pmse{0.6} & 16.5\pmse{0.9} & 20.6\pmse{1.2} & 25.8\pmse{1.6} & 23 \\
\actobs{}$\rightarrow$ECHO & 16 & 11.4\pmse{0.5} & 15.1\pmse{0.6} & 18.6\pmse{0.8} & 22.0\pmse{0.9} & 24.7\pmse{1.2} & 22 \\
$\lambda=0.05\rightarrow$GRPO & 8 & 11.4\pmse{0.7} & 15.2\pmse{0.8} & 18.8\pmse{1.0} & 21.3\pmse{1.1} & -- & -- \\
$\lambda=0.25\rightarrow$GRPO & 8 & 11.0\pmse{0.8} & \textbf{15.9}\pmse{0.9} & \textbf{20.1}\pmse{1.0} & 22.5\pmse{1.0} & -- & -- \\
\actobs{}-100$\rightarrow$GRPO & 8 & 11.0\pmse{0.6} & 14.0\pmse{0.7} & 16.5\pmse{0.9} & 19.1\pmse{1.1} & -- & -- \\
\bottomrule
\end{tabular}
\end{center}
\end{table}

Table~\ref{tab:full8b} contains the primary comparison together with
the diagnostic variants. \emph{act2act} repeats action-only SFT for a second
epoch. \emph{ActObs-100} applies joint action-observation supervision for the
first 100 steps and then continues with action-only SFT. \emph{Shuffled obs}
pairs actions with observations sampled from other trajectories, whereas
\emph{wild logs} replaces observations with token- and format-matched
production logs unrelated to the preceding action. The \emph{salient} variants
remove command echoes, repeated lines, and regular scaffold text before
observation supervision.

The primary rows show how the 8B ordering changes with sampling budget.
Before RL, \actsft{} has the highest point estimate at pass@1, while the
three principal SFT methods remain within a narrow band over the full
\passk{} range. After the same GRPO procedure, \actobs{} matches the leading
pass@4 score and becomes strongest at larger budgets, reaching 23.6 at
pass@8 and 27.0 at pass@16. It also solves 24 of 89 tasks, compared with 21
for \actsft{}$\rightarrow$GRPO. The sequential \obsact{} method receives
both forms of supervision but does not reproduce this high-budget advantage,
supporting joint rather than temporally separated supervision.

The remaining rows identify what produces that separation. A second epoch
of action-only SFT closely tracks \actsft{}, so additional optimization alone
does not recover the ActObs behavior. ActObs-100 reaches 19.1 pass@8 after
GRPO, below the 23.6 obtained when observation supervision continues through
SFT, indicating that the observation objective remains useful beyond the
early updates. Observation-only SFT and action SFT followed by
observation-only training solve no tasks, showing that observation prediction
complements action learning rather than replacing it. Correct pairing also
matters: shuffled observations reduce pass@1 from 7.9 to 3.7 and pass@4 from
16.1 to 9.0, whereas unrelated but clearly distinguishable wild logs retain
7.9 pass@1. Finally, supervision on the complete terminal response is stronger
than selecting only salient lines, consistent with the payload-level effects
in Figure~\ref{fig:position}(b,c).

\paragraph{Complete 4B matrix.}
\begin{table}[t]
\caption{\textbf{Complete Qwen3-4B results on \tbtwo{}.} Rows above the horizontal rule report SFT checkpoints, 
and rows below it
report RL checkpoints. Each checkpoint is evaluated with 16 attempts per
task.}
\label{tab:full4b}
\begin{center}
\footnotesize
\begin{tabular}{l ccccc c}
\toprule
Model & pass@1 & pass@2 & pass@4 & pass@8 & pass@16 & solved/89 \\
\midrule
Base (no SFT) & 1.2\pmse{0.2} & 2.1\pmse{0.4} & 3.3\pmse{0.5} & 4.7\pmse{0.6} & 5.6\pmse{0.7} & 5 \\
\actsft{} & \textbf{4.5}\pmse{0.4} & \textbf{7.0}\pmse{0.6} & \textbf{10.4}\pmse{0.8} & 14.0\pmse{1.0} & 16.9\pmse{1.2} & 15 \\
\actobs{} & 4.4\pmse{0.4} & 6.8\pmse{0.6} & 9.9\pmse{0.9} & 14.0\pmse{1.1} & \textbf{18.0}\pmse{1.4} & \textbf{16} \\
\obsact{} & 4.4\pmse{0.4} & 6.8\pmse{0.6} & 10.1\pmse{0.9} & \textbf{14.2}\pmse{1.1} & \textbf{18.0}\pmse{1.4} & \textbf{16} \\
\midrule
\actsft{}$\rightarrow$GRPO & 5.6\pmse{0.4} & 8.3\pmse{0.5} & 11.1\pmse{0.7} & 14.0\pmse{1.0} & 18.0\pmse{1.4} & 16 \\
\actobs{}$\rightarrow$GRPO & \textbf{7.2}\pmse{0.4} & \textbf{9.7}\pmse{0.5} & 12.5\pmse{0.7} & 15.5\pmse{1.0} & 19.1\pmse{1.4} & 17 \\
\obsact{}$\rightarrow$GRPO & 5.3\pmse{0.4} & 8.1\pmse{0.6} & 11.4\pmse{0.8} & 14.7\pmse{0.9} & 16.9\pmse{1.0} & 15 \\
\actsft{}$\rightarrow$ECHO & 5.9\pmse{0.4} & 8.6\pmse{0.6} & 11.6\pmse{0.8} & 14.7\pmse{1.0} & 18.0\pmse{1.3} & 16 \\
\actobs{}$\rightarrow$ECHO & 5.9\pmse{0.4} & 9.0\pmse{0.6} & \textbf{12.7}\pmse{0.8} & \textbf{16.5}\pmse{1.1} & \textbf{20.2}\pmse{1.4} & \textbf{18} \\
\bottomrule
\end{tabular}
\end{center}
\end{table}

\begin{table}[t]
\caption{\textbf{Direct \actobs{} versus \actsft{} comparisons.} Each cell
reports the \actobs{} minus \actsft{} difference on \tbtwo{} in
\passk{} points. The ECHO rows compare the corresponding ECHO checkpoints.}
\label{tab:direct}
\begin{center}
\footnotesize
\begin{tabular}{l ccccc}
\toprule
Comparison & $\Delta$ pass@1 & $\Delta$ pass@2 & $\Delta$ pass@4 & $\Delta$ pass@8 & $\Delta$ pass@16 \\
\midrule
4B after SFT
& -0.1 & -0.2 & -0.4 & 0.0 & +1.1 \\
4B after GRPO
& +1.5 & +1.5 & +1.4 & +1.4 & +1.1 \\
4B after ECHO
& 0.0 & +0.4 & +1.1 & +1.8 & +2.2 \\
\midrule
8B after SFT
& -1.3 & -1.8 & -2.0 & -2.2 & -2.2 \\
8B after GRPO
& -1.3 & -0.8 & +0.7 & +2.7 & +3.4 \\
8B after ECHO
& +1.8 & +2.2 & +2.2 & +1.4 & -1.1 \\
\bottomrule
\end{tabular}
\end{center}
\end{table}

Table~\ref{tab:full4b} repeats the full primary comparison at the smaller
model scale. It provides a stricter test of whether the downstream effect is
specific to one model size or to the 8B crossover pattern.

The 4B SFT checkpoints are almost indistinguishable at pass@1, with point
estimates between 4.4 and 4.5. Observation-supervised SFT nevertheless reaches
18.0 at pass@16 and solves one additional task. The difference becomes much
clearer after optimization: among the three plain-GRPO policies,
\actobs{}$\rightarrow$GRPO is strongest at every reported sampling budget. It
improves on \actsft{}$\rightarrow$GRPO by 1.5 points at pass@1 and retains an
advantage through pass@16. Thus, the 4B result expresses the same initialization
effect without requiring a crossover in the final \passk{} ordering.

The ECHO rows provide a complementary RL objective. With the ActObs
initialization, ECHO reaches 20.2 pass@16 and solves 18 tasks, the largest
high-budget point estimate and task count in the 4B matrix. Across both GRPO
and ECHO, the observation-aware initialization therefore supports stronger
\passk{} performance as the sampling budget grows.

\paragraph{Direct endpoint differences.}

Table~\ref{tab:direct} subtracts the \actsft{} score from the \actobs{} score
at matched model scale, training stage, and sampling budget. This presentation
removes the common performance level and exposes how the gap changes during RL.

At 4B, a near-zero SFT difference becomes a positive post-GRPO difference at
every $k$, with gains of 1.1--1.5 points. At 8B, ActObs begins below ActionSFT
at the SFT checkpoint, but GRPO shifts the comparison steadily toward ActObs
as the sampling budget grows: the difference crosses zero between pass@2 and
pass@4 and reaches +3.4 points at pass@16. The direction of change at both
scales shows that the central effect is produced by how RL uses the
initialization, rather than by a uniformly stronger SFT checkpoint.

The ECHO comparison is also informative. At 4B, its ActObs advantage grows
from parity at pass@1 to +2.2 points at pass@16. At 8B, the advantage is
concentrated at smaller and intermediate budgets. This distinction reinforces
that observation supervision changes the distribution of successful behavior,
while the precise conversion of that distribution depends on the downstream
RL objective.

\paragraph{Improvement during RL.}

\begin{table}[t]
\caption{\textbf{Performance gains during RL.} Each cell reports the change in
\passk{} relative to the corresponding SFT initialization, in points. Every checkpoint is evaluated with 16
attempts per task.}
\label{tab:gains}
\begin{center}
\footnotesize
\begin{tabular}{ll ccccc}
\toprule
& Init & $\Delta$ pass@1 & $\Delta$ pass@2 & $\Delta$ pass@4 & $\Delta$ pass@8 & $\Delta$ pass@16 \\
\midrule
\multirow{5}{*}{8B}
& \actsft{} & +3.1 & +2.2 & +0.5 & -1.0 & -1.1 \\
& \actobs{} & +3.1 & +3.2 & +3.2 & \textbf{+3.9} & \textbf{+4.5} \\
& \obsact{} & \textbf{+3.6} & \textbf{+3.8} & \textbf{+3.4} & +1.9 & -1.1 \\
& \actsft{} (ECHO) & +0.4 & -0.7 & -1.7 & -1.4 & +1.1 \\
& \actobs{} (ECHO) & \textbf{+3.6} & +3.3 & +2.5 & +2.2 & +2.2 \\
\midrule
\multirow{5}{*}{4B}
& \actsft{} & +1.1 & +1.3 & +0.8 & +0.1 & +1.1 \\
& \actobs{} & \textbf{+2.7} & \textbf{+3.0} & +2.6 & +1.5 & +1.1 \\
& \obsact{} & +0.8 & +1.3 & +1.3 & +0.5 & -1.1 \\
& \actsft{} (ECHO) & +1.4 & +1.6 & +1.2 & +0.7 & +1.1 \\
& \actobs{} (ECHO) & +1.5 & +2.2 & \textbf{+2.7} & \textbf{+2.5} & \textbf{+2.2} \\
\bottomrule
\end{tabular}
\end{center}
\end{table}

Final-policy differences combine two quantities: where each method starts and
how much it learns during RL. Table~\ref{tab:gains} isolates the second by
subtracting each method's own SFT score from its corresponding RL score.

The 8B contrast is sharpest in this view. GRPO from \actsft{} gains 3.1
points at pass@1, but the gain contracts at larger $k$ and is negative at
pass@8 and pass@16. GRPO from \actobs{} gains at least 3.1 points at every
budget, with the improvement increasing to 4.5 points at pass@16. Sequential
\obsact{} supervision produces strong gains at small $k$ but does not retain
them at pass@16. Joint supervision is therefore the only 8B initialization
whose GRPO improvement remains positive and substantial throughout the
reported sampling range.

At 4B, \actobs{} likewise produces the largest plain-GRPO gain from pass@1
through pass@8. Its +2.7-point pass@1 gain is more than twice the +1.1 points
from \actsft{}, while its pass@8 gain is +1.5 rather than +0.1. The ActObs
ECHO row also remains positive across the full range and is strongest at
intermediate and large budgets. Together, these rows show that observation
supervision improves the amount and breadth of performance that downstream
optimization can extract from an SFT checkpoint.

\paragraph{Observation-loss weight.}

\begin{table}[t]
\caption{\textbf{Observation-loss weight sweep at 8B.} The SFT block reports checkpoints
trained with different values of $\lambda$, together with the change in the
command-token top1-to-top2 margin relative to \actsft{} on a fixed trace set.
Intermediate values use four attempts per task; the $\lambda=0$ and
$\lambda=1$ endpoints use the \actsft{} and \actobs{} evaluations with
16 attempts per task. The GRPO block reports the resulting policies after the
common RL recipe, using eight attempts per task for
$\lambda\in\{0.05,0.25\}$ and 16 attempts per task for the endpoints.}
\label{tab:lambda}
\begin{center}
\footnotesize
\begin{tabular}{l cccc}
\toprule
\multicolumn{5}{l}{\emph{After SFT}} \\
$\lambda$ & pass@1 & pass@2 & pass@4 & $\Delta$ cmd margin \\
\midrule
0 (\actsft{}) & 9.2\pmse{0.5} & 13.6\pmse{0.7} & \textbf{18.2}\pmse{0.8} & -- \\
0.05 & 9.0\pmse{0.5} & 13.3\pmse{0.7} & 18.0\pmse{0.8} & -0.29 \\
0.25 & \textbf{9.3}\pmse{0.6} & \textbf{13.9}\pmse{0.7} & 18.0\pmse{0.8} & -0.35 \\
0.5 & 7.9\pmse{0.7} & 12.0\pmse{0.8} & 16.9\pmse{0.9} & -0.36 \\
1 (\actobs{}) & 7.9\pmse{0.5} & 11.8\pmse{0.7} & 16.1\pmse{0.8} & -0.48 \\
2.0 & 8.1\pmse{0.5} & 11.4\pmse{0.6} & 14.6\pmse{0.8} & -0.52 \\
\midrule
\multicolumn{5}{l}{\emph{After GRPO}} \\
$\lambda$ & pass@1 & pass@2 & pass@4 & pass@8 \\
\midrule
0 (\actsft{}) & \textbf{12.3}\pmse{0.5} & \textbf{15.9}\pmse{0.5} & 18.7\pmse{0.6} & 21.0\pmse{0.8} \\
0.05 & 11.4\pmse{0.7} & 15.2\pmse{0.8} & 18.8\pmse{1.0} & 21.3\pmse{1.1} \\
0.25 & 11.0\pmse{0.8} & \textbf{15.9}\pmse{0.9} & \textbf{20.1}\pmse{1.0} & 22.5\pmse{1.0} \\
1 (\actobs{}) & 11.0\pmse{0.5} & 15.0\pmse{0.7} & 19.3\pmse{0.9} & \textbf{23.6}\pmse{1.1} \\
\bottomrule
\end{tabular}
\end{center}
\end{table}

The binary comparison between ActionSFT and ActObs corresponds to the
endpoints $\lambda=0$ and $\lambda=1$ in Eq.~\ref{eq:actobs}.
Table~\ref{tab:lambda} adds intermediate weights to test whether the final
behavior changes gradually with the amount of observation supervision.

The SFT block shows that small observation weights preserve the immediate
benchmark level: $\lambda=0.05$ and $\lambda=0.25$ remain close to
ActionSFT through pass@4. At the same time, the command-token top-1 to top-2
margin decreases progressively as $\lambda$ grows, showing that the action
distribution changes even when SFT benchmark scores remain similar. After
GRPO, pass@8 increases monotonically from 21.0 at $\lambda=0$ to 21.3, 22.5,
and 23.6 as the observation weight rises. Pass@1 moves from 12.3 to 11.0 over
the same sweep. Observation weight therefore controls a tradeoff between
single-attempt reliability and multi-attempt task reach: lower weight favors
one attempt, while higher weight allocates more probability across behaviors
that become useful with multiple samples.

The intermediate checkpoints are important because they rule out an
all-or-nothing endpoint effect. In particular, $\lambda=0.25$ reaches 20.1
pass@4 and 22.5 pass@8 after GRPO while starting from an SFT checkpoint whose
pass@1 and pass@4 closely match ActionSFT. The downstream change is therefore
already visible before the full ActObs weight is applied.

\paragraph{Absolute performance across sampling budgets.}
\begin{figure}[t]
\centering
\includegraphics[width=\linewidth]{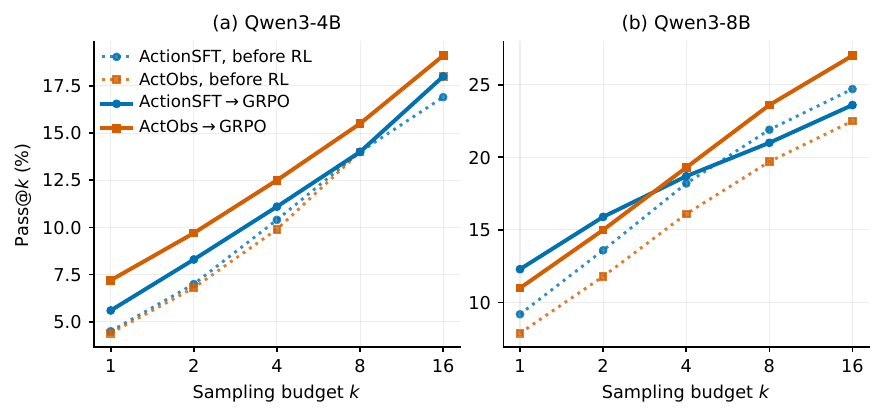}
\caption{\textbf{Absolute \passk{} curves.} The primary comparison at both
model scales complements Figure~\ref{fig:delta}. At 4B, the point estimate for
\actobs{}$\rightarrow$GRPO exceeds that for
\actsft{}$\rightarrow$GRPO at every $k$. At 8B, the point-estimate curves
cross between $k=2$ and $k=4$.}
\label{fig:absk}
\end{figure}

Figure~\ref{fig:absk} complements the improvement plot in
Figure~\ref{fig:delta} by showing the absolute scores of the four primary SFT
and GRPO policies. It makes the scale-dependent pattern easy to compare: the
4B ActObs GRPO policy is stronger at every evaluated $k$, whereas the 8B
curves cross between $k=2$ and $k=4$ before the ActObs advantage expands at
larger budgets.

Taken together, the control ladder isolates three ingredients. Action
competence must be maintained, mismatched observations are harmful, and the
two objectives are most effective when optimized jointly throughout SFT. The
weight sweep then shows that this effect is graded rather than binary. These
findings connect the complete result matrices to the
mechanism in Sections~\ref{sec:dynamics}--\ref{sec:where}: observation
supervision changes the checkpoint in ways that are only partially reflected
by its immediate SFT score but become consequential during RL and repeated
sampling.
\label{sec:}

\begin{figure}[t]
\centering
\includegraphics[width=\linewidth]{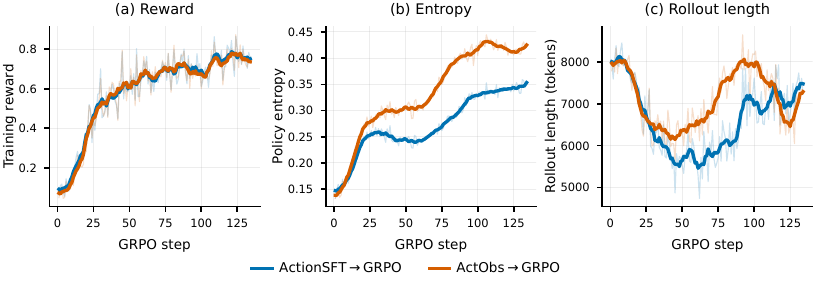}
\caption{\textbf{GRPO training dynamics at 8B.} This is the counterpart of
Figure~\ref{fig:dynamics}: (a) training reward, (b) on-policy training
entropy, and (c) rollout length. The two methods reach the same reward while
following clearly separated entropy trajectories. Both produce shorter
rollouts early in training; \actobs{} remains longer through the middle of
training, and the two methods end at similar lengths.}
\label{fig:dynamics8b}
\end{figure}

\begin{figure}[t]
\centering
\includegraphics[width=\linewidth]{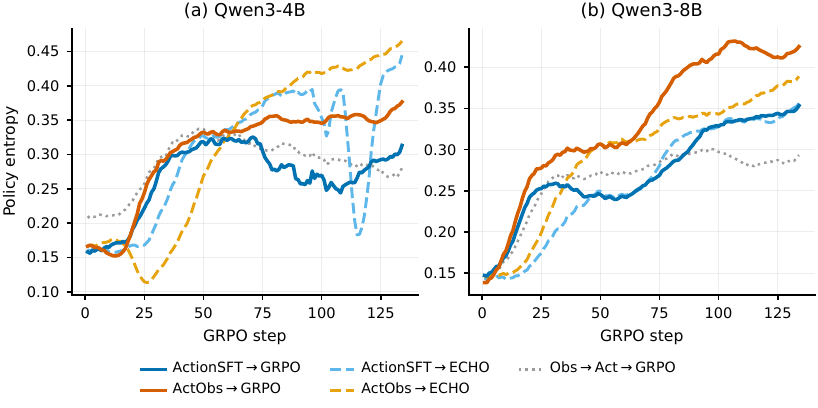}
\caption{\textbf{Training entropy across RL methods and model scales.}
Seven-step moving averages are shown for (a) Qwen3-4B and (b) Qwen3-8B.
Solid lines denote standard GRPO from \actsft{} or \actobs{}, dashed lines
denote ECHO, and the dotted line denotes standard GRPO from the sequential
\obsact{} initialization.}
\label{fig:entropy8b}
\end{figure}

\section{RL training dynamics}
\label{app:dynamics}
Figures~\ref{fig:dynamics8b} and~\ref{fig:entropy8b} extend the training-dynamics
analysis in Section~\ref{sec:dynamics}. Both figures report \emph{training
entropy}, the token-level policy entropy on the newly sampled on-policy batch
at each RL update. This differs from the endpoint self-entropy in
Figure~\ref{fig:mechlam}(a) and the fixed-state entropy in
Appendix~\ref{app:onpolicy}.

Figure~\ref{fig:dynamics8b} isolates the default GRPO comparison at 8B.
The reward curves in Figure~\ref{fig:dynamics8b}(a) overlap and reach the same
endpoint, but the entropy curves in Figure~\ref{fig:dynamics8b}(b) separate
near step 20 and remain apart. Figure~\ref{fig:dynamics8b}(c) rules out rollout
length as a simple explanation: \actobs{} is longer through the middle of
training, yet the two methods finish at similar lengths while their entropy
gap remains. Together with the shorter final ActObs rollouts at 4B, this shows
that its higher late-stage entropy is not a mechanical consequence of longer
generations.

Figure~\ref{fig:entropy8b} broadens the comparison to all five RL methods at
both model scales. At 4B, Figure~\ref{fig:entropy8b}(a) shows \actsft{} losing
entropy late in standard GRPO while \actobs{} remains higher. At 8B,
Figure~\ref{fig:entropy8b}(b) shows the ActObs separation emerging earlier and
growing through training. The sequential \obsact{} method finishes below both
at each scale, so observation exposure alone does not reproduce joint
supervision.

The ECHO trajectories are less monotonic and finish with high entropy, but the
benchmark ordering is not determined by entropy magnitude alone. At 4B,
\actsft{}$\rightarrow$ECHO and \actsft{}$\rightarrow$GRPO attain the same
pass@16 and solve the same number of tasks. At 8B,
\actobs{}$\rightarrow$ECHO has the second-highest training-entropy endpoint but
solves 22 tasks, compared with 24 for \actobs{}$\rightarrow$GRPO
(Table~\ref{tab:full8b}). Meanwhile, the 8B methods finish with closely matched
reward, and pass@16 on the RL training tasks saturates near 0.9. The persistent
entropy separation therefore reflects different optimization trajectories
that aggregate reward does not reveal.

\section{Entropy and command-margin probes}
\label{app:onpolicy}

\begin{table}[t]
\caption{\textbf{Entropy and command margins at 8B.} Each checkpoint scores
200 of its own \tbtwo{} evaluation rollouts (self) and a shared set of 200
traces (fixed). Command entropy and margin are computed at emitted command
tokens; a lower top-1 to top-2 margin indicates a flatter distribution.
Within each stage, boldface marks the highest entropy and lowest margin.}
\label{tab:sf}
\begin{center}
\footnotesize
\begin{tabular}{l cccc}
\toprule
Model & self-entropy & fixed-state entropy & cmd entropy & cmd margin \\
\midrule
\actsft{} & 0.683 & 0.683 & 0.308 & 7.00 \\
\actobs{} & \textbf{0.706} & 0.708 & 0.321 & \textbf{6.56} \\
\obsact{} & 0.704 & 0.711 & \textbf{0.328} & 6.65 \\
act2act & 0.678 & 0.678 & 0.296 & 7.05 \\
Shuffled obs & 0.702 & \textbf{0.728} & 0.317 & 6.59 \\
Wild logs & 0.654 & 0.693 & 0.285 & 7.01 \\
\midrule
\actsft{}$\rightarrow$GRPO & 0.716 & 0.719 & 0.312 & 7.55 \\
\actobs{}$\rightarrow$GRPO & 0.779 & 0.755 & 0.320 & 7.24 \\
\obsact{}$\rightarrow$GRPO & 0.684 & 0.717 & 0.259 & 7.78 \\
\actsft{}$\rightarrow$ECHO & 0.794 & \textbf{0.788} & 0.355 & 7.21 \\
\actobs{}$\rightarrow$ECHO & \textbf{0.804} & 0.784 & \textbf{0.363} & \textbf{6.44} \\
\bottomrule
\end{tabular}
\end{center}
\end{table}

\paragraph{Probe construction.}
Table~\ref{tab:sf} reports student-forcing measurements for the 8B
checkpoints. Self-entropy is computed on 200 \tbtwo{} rollouts generated by
each checkpoint, covering 81 to 87 tasks. Fixed-state entropy instead scores
every checkpoint on the same 200 \actsft{} traces. Command entropy and margin
restrict the calculation to emitted commands; a smaller margin places the
second-most likely token closer to the top choice.

\paragraph{Before RL.}
ActObs is consistently flatter than ActionSFT. Its self-entropy rises from
0.683 to 0.706, and the same ordering holds on fixed states, 
showing that the difference is not only due to state visitation. Its
command margin also falls from 7.00 to 6.56. The sequential \obsact{}
checkpoint is similarly flat before RL but does not retain this structure
after GRPO. The controls sharpen this interpretation: a second action-only
epoch makes the policy sharper; shuffled observations reproduce the ActObs
entropy and margin signature but perform much worse; and wild logs reproduce
neither effect. Flatter command distributions are therefore a consequence of
observation supervision, while strong performance also requires pairing
actions with their actual feedback.

\paragraph{After RL.}
The post-GRPO rows reproduce the training-entropy ordering from
Appendix~\ref{app:dynamics}. ActObs retains substantially more self-entropy
than ActionSFT, 0.779 versus 0.716. The advantage remains on the shared trace
set, 0.755 versus 0.719, and its command margin remains flatter, 7.24 versus
7.55. The sequential policy moves in the opposite direction, ending with the
lowest self-entropy and command entropy among the standard-GRPO policies.
Thus, the SFT similarity between ActObs and \obsact{} breaks decisively during
GRPO.
ECHO raises entropy further: \actobs{}$\rightarrow$ECHO has the highest
self-entropy and command entropy in the table and the lowest margin, yet it
trails \actobs{}$\rightarrow$GRPO at pass@16 and solves fewer tasks
(Table~\ref{tab:full8b}). High entropy and a flat margin can support
exploration, but their magnitude alone does not determine success.

\section{Extended-budget evaluation}
\label{app:timeout}

\begin{table}[t]
\caption{\textbf{Pass@1 after doubling the \tbtwo{} agent wall-clock budget.} The
extended-budget evaluation uses four attempts per task; the standard-budget
column reports the 16-attempt headline results. Every observation-supervised
checkpoint improves under the longer interaction budget.}
\label{tab:budget2x}
\begin{center}
\footnotesize
\begin{tabular}{l ccc}
\toprule
Model & pass@1 at 1$\times$ & pass@1 at 2$\times$ & $\Delta$ \\
\midrule
\actsft{} & 9.2 & 7.9 & -1.3 \\
\actobs{} & 7.9 & 10.1 & +2.2 \\
\obsact{} & 8.3 & 9.8 & +1.5 \\
\actsft{}$\rightarrow$GRPO & 12.3 & 14.0 & +1.8 \\
\actobs{}$\rightarrow$GRPO & 11.0 & 12.6 & +1.7 \\
\obsact{}$\rightarrow$GRPO & 11.9 & 12.1 & +0.2 \\
\bottomrule
\end{tabular}
\end{center}
\end{table}

We test whether observation-supervised initializations make productive use of
longer interaction horizons within \tbtwo{} by reevaluating six 8B checkpoints
after doubling the agent
wall-clock limit while leaving the verifier timeout and serving configuration
unchanged. Table~\ref{tab:budget2x} compares these four-attempt evaluations
with the headline results under the standard budget.

All four observation-supervised checkpoints have higher pass@1 point
estimates under the doubled budget. The SFT gains are +2.2 points for
\actobs{} and +1.5 for \obsact{}, and the post-GRPO gains are +1.7 and
+0.2 points. The standard \actsft{} checkpoint is the only SFT policy that
does not benefit from the additional interaction time. This consistent
directional pattern shows that policies initialized with observation
supervision continue to convert a longer interaction budget into successful
trajectories.

\section{Behavioral studies}
\label{app:cases}

\paragraph{Within-task strategy diversity.}
We conducted three blinded LLM-judge analyses with two judges per unit and
the method identity concealed. The judges counted distinct solution
strategies among the successful attempts for each task. Exact inter-judge
agreement ranges from 91\% to 94\% across the three analyses. In the largest
analysis, the methods average 1.5--1.9 strategies per solved task, and their
pairwise command-set Jaccard similarities are also closely matched. These
probes localize the ActObs advantage to \emph{outcome diversity}: its
successes are spread across more tasks, rather than repeating more distinct
strategies on tasks that are already solved.

\begin{table}[t]
\caption{\textbf{Success distribution across tasks.} Results are computed on
\tbtwo{} using 16 attempts per task for each checkpoint. Win entropy is the Shannon
entropy of the checkpoint's success distribution over its solved tasks; higher
values indicate more evenly spread successes. Boldface marks the largest value
within each stage and column.}
\label{tab:spread}
\begin{center}
\footnotesize
\begin{tabular}{l cccc}
\toprule
Model & Successes & Tasks solved & Successes/task & Win entropy (bits) \\
\midrule
\actsft{} & \textbf{131} & \textbf{22} & \textbf{5.95} & \textbf{4.11} \\
\actobs{} & 112 & 20 & 5.60 & 3.97 \\
\obsact{} & 118 & \textbf{22} & 5.36 & 3.94 \\
\midrule
\actsft{}$\rightarrow$GRPO & \textbf{175} & 21 & \textbf{8.33} & 4.05 \\
\actobs{}$\rightarrow$GRPO & 156 & \textbf{24} & 6.50 & \textbf{4.16} \\
\obsact{}$\rightarrow$GRPO & 169 & 21 & 8.05 & 4.10 \\
\actsft{}$\rightarrow$ECHO & 137 & 23 & 5.96 & 3.93 \\
\actobs{}$\rightarrow$ECHO & 163 & 22 & 7.41 & 4.06 \\
\bottomrule
\end{tabular}
\end{center}
\end{table}

\paragraph{Composition of the task-coverage difference.}
At 8B, \actobs{}$\rightarrow$GRPO and \actsft{}$\rightarrow$GRPO solve 19
tasks in common. ActObs solves five additional tasks:
\emph{code-from-image}, \emph{fix-git}, \emph{kv-store-grpc},
\emph{mcmc-sampling-stan}, and \emph{qemu-startup}. ActionSFT solves two
different tasks, giving ActObs a net advantage of three tasks, 24 versus 21.
Three of the five ActObs-only tasks, \emph{code-from-image},
\emph{kv-store-grpc}, and \emph{qemu-startup}, receive no success across the
64 combined SFT attempts per task or the 16 \actsft{}$\rightarrow$GRPO
attempts. Blinded review identifies two recurring patterns: less correlated
execution of a shared procedure across attempts and broader sampling of
parameters within a trajectory.

The per-task ledger supports the same picture. Relative to its SFT
initialization, \actobs{}$\rightarrow$GRPO adds six tasks while losing only
two; \actsft{}$\rightarrow$GRPO adds three and loses four. Across
\actsft{}$\rightarrow$GRPO, \actobs{}$\rightarrow$GRPO,
\obsact{}$\rightarrow$GRPO, and \actsft{}$\rightarrow$ECHO, 31 of 89 tasks
are solved by at least one policy. Table~\ref{tab:spread} shows how successes
are distributed over those tasks. At the SFT stage, all three checkpoints
average 5.4--6.0 successes per solved task. After RL,
\actsft{}$\rightarrow$GRPO concentrates 8.3 successes per solved task over
21 tasks, whereas \actobs{}$\rightarrow$GRPO distributes 6.5 successes per
task over 24 tasks and achieves the highest win entropy. This broader success
portfolio is the counterpart of its stronger high-$k$ performance.

\begin{table}[t]
\caption{\textbf{Behavioral patterns in post-RL trajectories.} Statistics
are computed over all available \tbtwo{} trajectories for each displayed
policy. Each value is paired with its policy; the counts in the final row are
totals within the two groups.}
\label{tab:traits}
\begin{center}
\footnotesize
\begin{tabular}{>{\raggedright\arraybackslash}p{2.6cm}
                >{\raggedright\arraybackslash}p{4.6cm}
                >{\raggedright\arraybackslash}p{4.6cm}}
\toprule
Measure & Higher self-entropy & Lower self-entropy \\
\midrule
Self-entropy & \actobs{}$\rightarrow$GRPO: 0.779;
\actsft{}$\rightarrow$ECHO: 0.794 &
\actsft{}$\rightarrow$GRPO: 0.716;
\obsact{}$\rightarrow$GRPO: 0.684 \\
Trajectory shape & \actobs{}$\rightarrow$GRPO uses 10.8 fewer commands per
attempt; \actsft{}$\rightarrow$ECHO uses 29.8 fewer. Both use more turns. &
More command-heavy repetition \\
Opening diversity & \actobs{}$\rightarrow$GRPO: 0.797;
\actsft{}$\rightarrow$ECHO: 0.772 &
\actsft{}$\rightarrow$GRPO: 0.760;
\obsact{}$\rightarrow$GRPO: 0.723 \\
Response when stalled & 42 outward escalations across both policies,
including new tools and orchestration scripts & 16 outward escalations
across both policies, with more terminal resets and repeated rewrites \\
\bottomrule
\end{tabular}
\end{center}
\end{table}

\paragraph{Case studies.}
We examine \emph{qemu-startup}, \emph{fix-git}, and \emph{kv-store-grpc},
three of the five tasks solved by \actobs{}$\rightarrow$GRPO but not by
\actsft{}$\rightarrow$GRPO.
Appendix~\ref{app:transcripts} reproduces trajectory excerpts for
\emph{qemu-startup} and \emph{fix-git}. On \emph{qemu-startup}, all methods use
the same general procedure: launch QEMU in the background with a serial TCP
server, then terminate and retry unsuccessful launches. The successful
\actobs{}$\rightarrow$GRPO trajectory systematically changes one launch
parameter at a time. In particular, it varies guest memory from
\texttt{-m 128M} to \texttt{512M} and then \texttt{1024M}; QEMU's 128 MiB
default is insufficient to boot the image. None of the lower-entropy
attempts samples this sequence.

On \emph{fix-git}, every method locates the dangling commit with
\texttt{git reflog}, but the successful ActObs trajectories continue trying
new integration tactics. The trajectory in Appendix~\ref{app:transcripts}
works through failed cherry-pick and patch attempts, reapplies the recovered
content, and commits it; another successful trajectory reaches the recovered
state with \texttt{git reset --hard}. Two ActionSFT trajectories reach the
integration step but subsequently lose progress because of pager-handling
failures and an unsupported file rewrite. On \emph{kv-store-grpc}, the methods
follow the same high-level procedure, but the ActObs attempts fail at less
correlated points and are more likely to complete the full sequence.

\paragraph{Behavioral patterns in post-RL trajectories.}
We analyze the available \tbtwo{} trajectories from four post-RL policies.
Based on the self-entropy measurements in Table~\ref{tab:sf}, we compare the
higher-entropy \actobs{}$\rightarrow$GRPO and \actsft{}$\rightarrow$ECHO
policies with the lower-entropy \actsft{}$\rightarrow$GRPO and
\obsact{}$\rightarrow$GRPO policies. The opening-diversity score measures
variation among the first three commands across attempts; a larger value
indicates less repetitive openings. Command-count differences are measured
relative to \actsft{}$\rightarrow$GRPO.

Table~\ref{tab:traits} shows a consistent behavioral separation. The
higher-entropy policies begin with more varied command sequences, issue fewer
commands across more turns, and more often change tactics when progress
stalls. The lower-entropy policies rely more heavily on dense command
sequences, terminal resets, and repeated rewrites. Opening diversity is a
correlate of entropy rather than the main source of success. In a blinded
review of 12 successful trajectories from tasks where the methods differ,
eight successes arise from less correlated executions of the same high-level
procedure, two from trying different parameter values, and two from sustained
persistence.

\begin{table}[t]
\caption{\textbf{Trajectory-level behavioral statistics.} Statistics are
computed over all available \tbtwo{} trajectories for each policy. ``Median
steps'' is the median number of interaction steps across all trajectories,
whereas ``Solve steps'' is the median restricted to successful trajectories.
``Repeats/traj.'' and ``Errors/traj.'' are the mean numbers of exact command
repetitions and error-bearing observations per trajectory, respectively.
``Retry-same'' is the fraction of error-bearing observations followed by an
exact repetition of the preceding command.}
\label{tab:battery}
\begin{center}
\footnotesize
\begin{tabular}{l ccccc}
\toprule
Model & Median steps & Solve steps & Repeats/traj. & Errors/traj. & Retry-same \\
\midrule
Base & 13 & 5 & 19.0 & 10.2 & 40.5\% \\
\actsft{} & 51 & 21 & 45.7 & 23.7 & 55.5\% \\
\actobs{} & 59 & 13 & 48.1 & 26.8 & 51.9\% \\
\obsact{} & 49 & 11 & 34.9 & 19.7 & 48.0\% \\
act2act & 41 & 14 & 45.8 & 22.2 & 53.4\% \\
Wild logs & 37 & 14 & 33.3 & 15.0 & 42.6\% \\
Shuffled obs & 55 & 17 & 52.9 & 33.4 & 75.9\% \\
\midrule
\actsft{}$\rightarrow$GRPO & 41 & 26 & 1.0 & 12.4 & 0.7\% \\
\actobs{}$\rightarrow$GRPO & 48 & 35 & 1.2 & 10.9 & 0.9\% \\
\obsact{}$\rightarrow$GRPO & 52 & 33 & 1.0 & 12.0 & 0.9\% \\
\actsft{}$\rightarrow$ECHO & 34 & 29 & 1.0 & 6.8 & 0.8\% \\
\actobs{}$\rightarrow$ECHO & 29 & 21 & 0.6 & 10.4 & 0.9\% \\
\bottomrule
\end{tabular}
\end{center}
\end{table}

\paragraph{Trajectory-level statistics.}

Table~\ref{tab:battery} separates overall trajectory length from the length
of successful trajectories. The two quantities capture different behaviors:
the first reflects how long a policy continues interacting across both solved
and unsolved attempts, while the second measures how directly its successful
attempts reach a solution. Before RL, \actobs{} has a longer median trajectory
than \actsft{}, 59 versus 51 steps, but its successful trajectories are much
shorter, 13 versus 21 steps. The sequential \obsact{} policy has the shortest
successful trajectories at 11 median steps. Observation-supervised SFT can
therefore combine greater persistence overall with more efficient execution
on the attempts that succeed.

The clearest shared effect of GRPO is a collapse in redundant command
repetition. Across the three main SFT-to-GRPO comparisons, exact repetitions
fall from 34.9--48.1 per trajectory to 1.0--1.2, and identical retries after
an error fall from 48.0--55.5\% to 0.7--0.9\%. Errors per trajectory also
decline for every main method. Overall trajectory length, however, does not
uniformly shrink: the ActObs and \obsact{} endpoints remain longer than the
ActionSFT endpoint. Their behavior is therefore better characterized by less
repetitive interaction than by shorter interaction. The shuffled-observation
control provides the opposite pattern. It has the highest error count and
repeats the preceding command after most errors, showing that observation
tokens help only when they are paired with the actions that produced them.

\section{Representative transcripts}
\label{app:transcripts}

This appendix reproduces excerpts from the \actobs{}$\rightarrow$GRPO
trajectories used in the \emph{qemu-startup} and \emph{fix-git} case studies
in Appendix~\ref{app:cases}. All text is quoted
verbatim from the recorded trajectories; \texttt{[...]}\ marks elided text,
and trailing blank screen lines are trimmed. Blue boxes show agent turns: the
\texttt{analysis} and \texttt{plan} fields of the JSON response, followed by
the submitted keystrokes. Orange boxes show the observation returned to the
model after execution. Long episodes also contain scaffold-managed context
summarization turns, which we elide.

We first show the episode prompt that defines the environment. Every
episode begins with a single user message containing the scaffold
instructions, the task description, and the initial terminal screen; each
subsequent user message contains only the new terminal output.

\begin{taskbox}[terminus-2 episode prompt (first user message, excerpt)]
\begin{Verbatim}
You are an AI assistant tasked with solving command-line tasks in a Linux
environment. You will be given a task description and the output from
previously executed commands. Your goal is to solve the task by providing
batches of shell commands.

Format your response as JSON with the following structure:

{
  "analysis": "Analyze the current state based on the terminal output
    provided. What do you see? What has been accomplished? What still
    needs to be done?",
  "plan": "Describe your plan for the next steps. [...]",
  "commands": [
    {
      "keystrokes": "ls -la\n",
      "duration": 0.1
    }
  ],
  "task_complete": true
}
[...]
IMPORTANT: The text inside "keystrokes" will be used completely verbatim
as keystrokes. Write commands exactly as you want them sent to the
terminal:
- You must end every command with a newline (\n) or it will not execute.
- For special key sequences, use tmux-style escape sequences:
  - C-c for Ctrl+C
[...]
Task Description:
<task text, see below>

Current terminal state:
Current Terminal Screen:
root@b6641dc52806:/app#
\end{Verbatim}
\end{taskbox}

\subsection{Case study: qemu-startup}
\label{app:transcript-qemu}

Among the evaluated policies, \actobs{}$\rightarrow$GRPO uniquely solves
this task. The decisive difference is guest memory: the other attempts use
the default 128\,MiB, too little for the image to produce a login prompt on
the serial port.

\begin{taskbox}[Task description]
\begin{Verbatim}
Start the /app/alpine.iso image in qemu in such a way that I can connect
to it via `telnet 127.0.0.1 6665`. When I run telnet I will expect to see
the login prompt; I'll log in. You should just get me to this point.
Start the image in the background and leave it running. Block until it's
ready.
\end{Verbatim}
\end{taskbox}

The winning attempt begins with the same canonical launch every method uses:

\begin{agentturn}[(early)]
\begin{Verbatim}
Analysis: File command not installed, but we can still start qemu. Need
to start the ISO image with serial console TCP on port 6665. We'll start
qemu in the background, then wait for it to be ready. [...]

Commands:
qemu-system-x86_64 -m 128M -cdrom /app/alpine.iso
  -serial tcp:127.0.0.1:6665,wait -nographic > /tmp/qemu.log 2>&1 &
\end{Verbatim}
\end{agentturn}

\begin{envturn}[(observation)]
\begin{Verbatim}
New Terminal Output:

root@b6641dc52806:/app# qemu-system-x86_64 -m 128M -cdrom /app/alpine.iso
-serial tcp:127.0.0.1:6665,wait -nographic > /tmp/qemu.log 2>&1 &
[1] 131
root@b6641dc52806:/app#
\end{Verbatim}
\end{envturn}

The VM never becomes reachable, and the log stays empty. Where the
low-entropy policies re-issue near-identical launches, this trajectory begins
a search that changes one launch parameter at a time, killing the previous
process between attempts. The session recording contains twelve distinct
QEMU invocations before the final one, varying the serial backend, boot
flags, drive interface, and, critically, guest memory:

\begin{taskbox}[QEMU launch variants attempted (from the session recording,
condensed; each launch is followed by a check and a kill)]
\begin{Verbatim}
-m 128M ... -serial tcp:127.0.0.1:6665,wait -nographic
-m 128M ... -serial tcp:127.0.0.1:6665,server,nowait -nographic
-m 128M ... -serial tcp:...,server,nowait -nographic -nodefaults
-m 128M ... -boot c -serial tcp:...,server,nowait -nographic
-m 128M -drive file=/app/alpine.iso,if=ide,format=raw -serial tcp:...
-m 512M ... -serial tcp:...,server,nowait -nographic -nodefaults
-m 1024M ... -serial tcp:...,server,nowait -nographic -nodefaults
-m 128M ... -serial stdio -nographic
-m 256M ... -boot d -serial tcp:...,server,nowait -nographic -nodefaults
-m 512M ... -boot d -serial tcp:...,server,nowait -nographic -nodefaults
-m 256M ... -serial stdio -nographic -nodefaults
-m 1024M ... -serial file:/tmp/serial_fifo -nographic -nodefaults
\end{Verbatim}
\end{taskbox}

\begin{agentturn}[(mid-episode, during the memory sweep)]
\begin{Verbatim}
Analysis: We have killed the latest qemu process. Need to start the ISO
with a serial console TCP server. The ISO is valid. The previous attempts
were killed due to high CPU, indicating the image might be stuck booting.
[...]

Commands:
qemu-system-x86_64 -m 512M -cdrom /app/alpine.iso
  -serial tcp:127.0.0.1:6665,server,nowait -nographic -nodefaults
  > /tmp/qemu9.log 2>&1 &
\end{Verbatim}
\end{agentturn}

Between launches the trajectory also runs a forensic detour, inspecting
the image with \texttt{od}, \texttt{file}, \texttt{xorriso}, and
\texttt{7z}, and attempting a FIFO-plus-\texttt{socat} serial relay. None
of these succeeds, and the agent returns to the direct launch, now with
the large memory setting:

\begin{agentturn}[(late)]
\begin{Verbatim}
Analysis: [...] We should now start QEMU with the correct command. We'll
also need to block until it's ready, meaning we need to wait for the
system to boot. [...]

Commands:
qemu-system-x86_64 -m 1024M -cdrom /app/alpine.iso
  -serial tcp:127.0.0.1:6665,server,nowait -nographic -nodefaults
  > /tmp/qemu.log 2>&1 &
\end{Verbatim}
\end{agentturn}

\begin{envturn}[(observation)]
\begin{Verbatim}
New Terminal Output:

root@b6641dc52806:/app# qemu-system-x86_64 -m 1024M -cdrom /app/alpine.iso
-serial tcp:127.0.0.1:6665,server,nowait -nographic -nodefaults >
/tmp/qemu.log 2>&1 &
[1] 2694
root@b6641dc52806:/app#
\end{Verbatim}
\end{envturn}

\begin{agentturn}[(verification)]
\begin{Verbatim}
Analysis: The QEMU process is running (PID 2694). The log file is empty,
indicating that QEMU hasn't produced any output yet. The system may still
be booting. We need to wait for the system to become ready (i.e., the
login prompt appears). [...]

Commands:
nc -z 127.0.0.1 6665 && echo 'Port 6665 is listening'
  || echo 'Port 6665 is not listening'
\end{Verbatim}
\end{agentturn}

\begin{envturn}[(observation)]
\begin{Verbatim}
New Terminal Output:

root@b6641dc52806:/app# nc -z 127.0.0.1 6665 && echo 'Port 6665 is
listening' || echo 'Port 6665 is not listening'
Port 6665 is listening
root@b6641dc52806:/app#
\end{Verbatim}
\end{envturn}

\begin{envturn}[(observation, after \texttt{telnet 127.0.0.1 6665})]
\begin{Verbatim}
New Terminal Output:

root@b6641dc52806:/app# telnet 127.0.0.1 6665 2>&1 | head -5
Trying 127.0.0.1...
Connected to 127.0.0.1.
Escape character is '^]'.
\end{Verbatim}
\end{envturn}

The trajectory then blocks, waiting for the guest to finish booting, and
is still waiting when the 900-second task wall-clock expires. The verifier
nevertheless finds the correctly configured VM running with its login
prompt reachable and scores the trial 1.0. The attempt therefore uses its
full interaction budget productively and leaves the environment in the
required state.

\subsection{Case study: fix-git}
\label{app:transcript-git}

\begin{taskbox}[Task description]
\begin{Verbatim}
I just made some changes to my personal site and checked out master, but
now I can't find those changes. Please help me find them and merge them
into master.

Current terminal state:
Current Terminal Screen:
root@03738d59a47f:/app/personal-site#
\end{Verbatim}
\end{taskbox}

Every method reaches the key discovery within a few turns; only
\actobs{}$\rightarrow$GRPO carries the integration through. The discovery:

\begin{agentturn}[(early)]
\begin{Verbatim}
Analysis: Only master branch exists. Need to find where the changes are.
Could be that the changes are in a commit that is not yet on master
(maybe we have a local branch that we haven't merged). [...] Let's
examine git reflog [...]

Commands:
git reflog
\end{Verbatim}
\end{agentturn}

\begin{envturn}[(observation)]
\begin{Verbatim}
New Terminal Output:

root@03738d59a47f:/app/personal-site# git reflog
d7d3e4b (HEAD -> master) HEAD@{0}: checkout: moving from
650dba427e0a9dcd118f41a4c5e35c8017550a5a to master
650dba4 HEAD@{1}: commit: Move to Stanford
c4e38a1 HEAD@{2}: checkout: moving from master to HEAD~1
d7d3e4b (HEAD -> master) HEAD@{3}: reset: moving to d7d3e4b
b0d59cd HEAD@{4}: clone: from
https://github.com/TheMikeMerrill/personal-site.git
root@03738d59a47f:/app/personal-site#
\end{Verbatim}
\end{envturn}

The dangling commit \texttt{650dba4} (``Move to Stanford'') is the lost
work. The integration is where the methods separate. A first
\texttt{git cherry-pick 650dba4} traps the session in a pager:

\begin{envturn}[(observation, after \texttt{git cherry-pick 650dba4})]
\begin{Verbatim}
Current Terminal Screen:
[...]
  -x [N[,...]]  --tabs=[N[,...]]
                  Set tab stops.
  -X  ........  --no-init
                  Don't use termcap init/deinit strings.
  -y [N]  ....  --max-forw-scroll=[N]
                  Forward scroll limit.
\end{Verbatim}
\end{envturn}

The trajectory escapes the pager and works through a sequence of failing
alternatives, reading each error and switching tactics rather than
re-issuing the same command:

\begin{envturn}[(observation, after \texttt{git cherry-pick -m 1 650dba4})]
\begin{Verbatim}
New Terminal Output:

root@03738d59a47f:/app/personal-site# git cherry-pick -m 1 650dba4
error: your local changes would be overwritten by cherry-pick.
hint: commit your changes or stash them to proceed.
fatal: cherry-pick failed
root@03738d59a47f:/app/personal-site#
\end{Verbatim}
\end{envturn}

\begin{envturn}[(observation, after \texttt{git apply -p1 <(git show
650dba4)})]
\begin{Verbatim}
New Terminal Output:

root@03738d59a47f:/app/personal-site# git apply -p1 <(git show 650dba4)
/dev/fd/63:17: trailing whitespace.
[...]
error: patch failed: _includes/about.md:1
error: _includes/about.md: patch does not apply
error: patch failed: _layouts/default.html:27
error: _layouts/default.html: patch does not apply
root@03738d59a47f:/app/personal-site#
\end{Verbatim}
\end{envturn}

Having already re-applied the commit's content to the working tree by
hand, the agent stages it and commits:

\begin{agentturn}[(final)]
\begin{Verbatim}
Analysis: The cherry-pick failed due to missing editor. We need to commit
the changes that are already staged. We can use git commit -m "Merge
changes from cherry-pick 650dba4". That should work because the changes
are already in the index. Let's try that.

Commands:
git commit -m "Merge changes from cherry-pick 650dba4"
\end{Verbatim}
\end{agentturn}

\begin{envturn}[(observation)]
\begin{Verbatim}
New Terminal Output:

root@03738d59a47f:/app/personal-site# git commit -m "Merge changes from
cherry-pick 650dba4"
[master ef1fb37] Merge changes from cherry-pick 650dba4
 Date: Sat Sep 13 17:54:33 2025 +0000
 2 files changed, 19 insertions(+), 17 deletions(-)
root@03738d59a47f:/app/personal-site#
\end{Verbatim}
\end{envturn}

The verifier scores the trial 1.0. For contrast, the
\actsft{}$\rightarrow$GRPO attempts on this task twice reach the same
integration step (\texttt{git show 650dba4} followed by a merge attempt)
and then lose the session to pager handling and an incorrect full-file
rewrite (Appendix~\ref{app:cases}). The difference in these transcripts is
not the initial plan, which is shared, but the ability to keep sampling new
tactics in response to observed errors until one succeeds.

\section{Limitations}
\label{app:limitations}

\paragraph{Domain coverage.}
Our experiments are confined to terminal agents. Aider-polyglot adds a
cross-domain evaluation on multilingual code editing. The results do not yet
establish whether observation supervision provides the same benefit for agents
operating in coding, web, GUI, or embodied environments. These settings also expose
richer observations than terminal text, making them an important direction for
testing the generality of the method.

\paragraph{Model and data scope.}
We study Qwen3 at 4B and 8B, which provides a cross-scale comparison but not a
cross-family one because the models share a tokenizer and chat template. All
SFT trajectories are generated by a single teacher, DeepSeek-V3.2, and contain
only the consequence realized after each teacher action.
Experiments with additional model families, teachers, and data collection 
pipelines would separate these factors.

\paragraph{Scaffold parity.}
The SFT, RL, and evaluation pipelines are not fully identical. They differ in
observation formatting and context construction, and RL uses a fresh,
non-interactive shell whereas \tbtwo{} uses a persistent \texttt{tmux} session.
Command handling, requested waits, generation templates, and termination and
scoring behavior also vary across pipelines. Because every method within a
stage uses the same scaffold, the central ActObs versus ActionSFT comparisons
remain controlled, although these mismatches may affect absolute scores and
comparisons between SFT and RL endpoints.

\section{Statements}

\paragraph{Author contributions.}

Juzheng Zhang identified the research direction, conceived the method, designed
and conducted the experiments, performed the analyses, and led the writing of
the manuscript.
Disha Makhija, Manoj Ghuhan
Arivazhagan, and Vinayshekhar Bannihatti
Kumar advised the project and contributed through technical discussions and
feedback. Rashmi Gangadharaiah led the team, supervised the project, and
provided strategic guidance. All authors contributed to writing the paper and
reviewed and approved the final manuscript.

\paragraph{AI use.}

Generative AI tools were used to assist with code development, language editing and polishing,
and the development of scripts for generating figures. The authors reviewed and
verified all AI-assisted outputs, including code, text, and scripts. All authors reviewed
and approved these uses and take full responsibility for the content of the final manuscript.

\paragraph{Reproducibility.}

Section~\ref{sec:setup} and Appendix~\ref{app:method} specify the corpus,
training recipes, serving configuration, evaluation convention, and
statistical procedures; Appendix~\ref{app:fulltable} reports the full method
matrix with per-method attempt counts. The loss-mask change is fully
specified by Eq.~\ref{eq:actobs}. Upon acceptance, we will release the code, data,
per-task evaluation outcomes, training and evaluation configurations, and analysis
artifacts needed to reproduce the reported results.

\end{document}